%% file: main.tex
\ifdefined\XeTeXversion\else\pdfoutput=1\fi
\documentclass[11pt]{article}

\usepackage{acl}
\usepackage{times}
\usepackage{latexsym}
\usepackage[T1]{fontenc}
\usepackage[utf8]{inputenc}
\usepackage{microtype}
\IfFileExists{inconsolata.sty}{\usepackage{inconsolata}}{}
\usepackage{amsmath}
\usepackage{amssymb}
\usepackage{booktabs}
\usepackage{multirow}
\usepackage{xcolor}
\usepackage{url}
\usepackage{graphicx}
\usepackage{placeins}
\usepackage{dblfloatfix}
\usepackage{balance}
\usepackage{algorithm}
\usepackage{algpseudocode}
\usepackage{amsthm}
\newtheorem{lemma}{Lemma}

\newcommand{\method}{IRDS}
\newcommand{\methodlong}{Interpretable RLVR Data Selection}
\newcommand{\modelname}[1]{\textnormal{\textsc{#1}}}
\newcommand{\datasetname}[1]{\textnormal{\textsc{#1}}}
\newcommand{\baseline}[1]{\textnormal{\textsc{#1}}}
\newcommand{\R}{\mathbb{R}}
\newcommand{\E}{\mathbb{E}}

\DeclareMathOperator{\diag}{diag}

\title{IRDS: Interpretable RLVR Data Selection via Verifier-Coupled Sparse Autoencoder Coverage}

\author{
  Yuhan Li\textsuperscript{1},
  Mingxu Zhang\textsuperscript{1},
  Dazhong Shen\textsuperscript{2}\thanks{Corresponding authors.},
  Ying Sun\textsuperscript{3}\footnotemark[1] \\
  \textsuperscript{1}The Hong Kong University of Science and Technology (Guangzhou) \\
  \textsuperscript{2}Nanjing University of Aeronautics and Astronautics \\
  \textsuperscript{3}The 63rd Research Institute, National University of Defense Technology, Nanjing \\
  \texttt{yuhanli530@gmail.com} \\
  \texttt{shendazhong@nuaa.edu.cn}, \texttt{sunyinggilly@gmail.com}
}


\begin{document}
\maketitle

\begin{abstract}
Reinforcement learning with verifiable rewards (RLVR) has become a key technique for enhancing LLM reasoning, yet its data inefficiency remains a major bottleneck. Existing methods address this problem only partially, each missing at least one of subset-level coverage, verifier signal use, or interpretability. To address this gap, we present \method{} (\methodlong{}), which selects RLVR training instances on a sparse autoencoder (SAE) cluster basis so the selection itself is auditable on recognizable problem motifs. To select instances the model both fails on and can still learn from, we introduce a verifier-coupled coverage objective on the SAE basis and solve it by greedy log-determinant maximization. Experiments on three instruction-tuned models and six math reasoning benchmarks show that \method{} achieves the highest overall accuracy, exceeding the strongest baseline by $+3.9/+4.0$\,pp on the two Qwen models and by $+0.5$\,pp on Llama-3.1-8B, while running an order of magnitude cheaper than the trajectory-based baseline.
\end{abstract}

\section{Introduction}

Reinforcement learning with verifiable rewards (RLVR) trains large language models using a rule-based verifier that scores each rollout as correct or incorrect, replacing the human preference labels used in earlier alignment pipelines. The recipe drives most recent progress on mathematical reasoning benchmarks such as MATH and AIME, where DeepSeekMath and DeepSeek-R1 together with their successors converged on it as the default approach to math post-training~\citep{shao2024deepseekmath,guo2025deepseekr1}. Under this recipe, every training instance contributes to the final policy through its sampled rollouts and the verifier's binary feedback, so the composition of the training set decides what the policy learns at each step. Three failure modes show up immediately in practice. An instance on which the policy already succeeds leaves nothing to repair; an instance on which it always fails produces no within-group reward variation; in both tails the group-relative gradient vanishes. A subset whose instances target overlapping weaknesses pays for the same correction many times. Selecting a subset that is simultaneously improvable, trainable, and diverse is therefore central to RLVR quality at a fixed data budget.

Several lines of work tackle this. Pointwise scalar methods~\citep{li2025limr,yi2026learnmoreless,zeng2025cures1037} rank instances by a single utility such as a difficulty estimate or a learning-impact trajectory and keep the top entries, but a sort on one number cannot tell when two top-ranked instances target the same policy weakness. Diversity-aware methods~\citep{yang2025saedataselection,ma2025taskspecific5573,tang2025towards1321} embed instances in a learned space and pick a spread-out subset, but the embedding is built without the verifier signal, so the diversity budget is spent on mathematical structure that the policy may already solve. Gradient-based methods~\citep{xia2024less,zhu2026data6491} prefer instances whose gradients align with a target signal, but the gradient lives in a dense parameter space whose directions resist human inspection. None of these lines jointly addresses failure relevance, trainability, and non-redundancy on an auditable basis.

We propose \method{} (\methodlong{}), an offline data-selection method that builds a verifier-coupled coverage objective on top of a sparse-autoencoder representation of the training set. The sparse-autoencoder basis gives a coordinate that is more semantic than surface features such as length or notation density, yet more inspectable than dense hidden states; each axis carries a recognizable problem motif such as geometry, divisor counting, or multiple-choice templates. On this coordinate, \method{} attaches two verifier-coupled weights per instance that disentangle which weakness still needs repair from how trainable the rollouts are, and combines the two into a verifier-coupled metric whose leading directions concentrate difficulty relative to trainability mass. A greedy log-determinant procedure on this metric returns a non-redundant subset that is simultaneously failure-relevant, trainable, and diverse. Across three instruction-tuned models and six math benchmarks under a shared GRPO recipe, \method{} attains the highest Overall accuracy at an order of magnitude lower cost than the trajectory-based baseline, while keeping the SAE block of the selection auditable.

Our contributions are as follows:
\begin{itemize}
\setlength{\itemsep}{1pt}
\item To our knowledge, \method{} is the first RLVR data-selection method built on a sparse-autoencoder cluster basis, which makes the selection auditable on recognizable problem motifs.
\item We design a verifier-coupled coverage objective on this basis that jointly targets failure-relevance, trainability, and non-redundancy, and solve it by greedy log-determinant maximization.
\item Experiments on three instruction-tuned models and six math benchmarks show that \method{} achieves the best Overall accuracy at every budget, with margins of $+3.9/+4.0$\,pp on the two Qwen models and $+0.5$\,pp on Llama-3.1-8B, at order-of-magnitude lower selection cost than the trajectory-based baseline.
\end{itemize}

\section{Related Work}
\label{sec:related}

\paragraph{Data Selection for RLVR.}
Recent RLVR data selection follows three lines, each addressing failure relevance, trainability, or coverage in isolation. The largest line of work attaches a scalar utility to each instance and keeps the top entries; utilities include difficulty estimates~\citep{gao2025prompt1135,zhao2026difficulty6375}, uncertainty scores~\citep{yi2026learnmoreless,zeng2025cures1037}, learning-impact trajectories~\citep{li2025limr}, and mid-difficulty rollout-replay~\citep{sun2025improving5316,fatemi2026prioritized2648}. A second line organizes instances on a learned embedding and picks a spread-out subset~\citep{tang2025towards1321,bhattacharyya2026finescope0624,zhou2026rethinking4981}, with the embedding built without the verifier signal. A third line aligns per-instance gradients with a target signal~\citep{xia2024less,zhu2026data6491}. \method{} differs from all three by representing instances on a sparse-autoencoder basis and lifting verifier-coupled coverage to the set level.

\paragraph{Sparse Autoencoders for Data Selection.}
Sparse autoencoders~\citep{bricken2023monosemanticity,cunningham2023sparse,gao2024scalingsaes,bussmann2024batchtopk} decompose neural activations into sparse codes over an overcomplete dictionary in which many directions correspond to recognizable concepts, which has made them a useful coordinate for downstream analysis. Two prior works use SAE features for data selection in instruction-tuning settings~\citep{yang2025saedataselection,ma2025taskspecific5573}; \method{} differs from both in three concrete ways: (i) a verifier-coupled \emph{failure} weight $d_i$ that drives the budget toward SAE directions where the policy still answers wrong, which neither prior method uses because instruction-tuning has no rollout-level correctness signal; (ii) a verifier-coupled \emph{trainability} weight $r_i$ that downweights the all-fail and all-solve tails where group-relative RLVR advantages vanish, which is specific to RL with verifiable rewards; and (iii) a \emph{set-level} coverage objective (greedy log-determinant on a whitened covariance) rather than the pointwise SAE-utility ranking used by both prior works, so redundancy between two equally hard-and-trainable instances is paid only once.

\section{Preliminaries}
\label{sec:problem-setup}

\paragraph{Verifiable-reward reinforcement learning.}
RLVR replaces the learned reward model of preference-based alignment with a programmatic verifier that returns a binary correctness label per rollout. Concretely, given a training set $\mathcal{D}=\{x_i\}_{i=1}^{N}$ and a model $\pi_\theta$, each training step draws an instance $x_i$, samples $G_i$ rollouts from $\pi_\theta(\cdot\mid x_i)$, and the verifier returns one bit $r_{ij}\!\in\!\{0,1\}$ per rollout. GRPO~\citep{shao2024deepseekmath}, the recipe we adopt throughout, then forms group-relative advantages from the $G_i$ rollouts of each instance and updates $\pi_\theta$ with a PPO-style clipped objective. Under group-relative objectives, instances whose rollouts all receive the same verifier label produce degenerate within-group advantages; for a Bernoulli success model with true success rate $P_i$, the response-level reward variance contains a $P_i(1{-}P_i)$ factor that vanishes at the all-fail and all-solve tails. In our experiments we sample a fixed $G_i\!=\!G\!=\!8$ rollouts per instance throughout, so the only signal consumed by selection is the per-instance success count $s_i=\sum_{j=1}^{G} r_{ij}\in\{0,1,\ldots,G\}$, an integer summary of the verifier's behavior on $x_i$ under $\pi_\theta$.

\paragraph{Sparse Autoencoders.}
A sparse autoencoder (SAE)~\citep{bricken2023monosemanticity,cunningham2023sparse,gao2024scalingsaes} decomposes an internal activation $h\in\R^{d_{\mathrm{model}}}$ into a sparse nonnegative code over an overcomplete dictionary of size $D\!\gg\!d_{\mathrm{model}}$, with a linear decoder reconstructing $\hat h = W_{\mathrm{dec}}\,a(h) + b_{\mathrm{dec}}$. We adopt the \emph{BatchTopK} variant~\citep{bussmann2024batchtopk} to let the per-token sparsity adapt to token-level information content while preserving an average sparsity of $k$ per token. For a batch $\mathcal{B}$ of $B$ tokens, the encoder first produces post-ReLU activations $\tilde z_t = \mathrm{ReLU}\!\bigl(W_{\mathrm{enc}}(h_t - b_{\mathrm{dec}}) + b_{\mathrm{enc}}\bigr)$, and BatchTopK then zeros all but the $Bk$ largest entries among $\{\tilde z_{t,i}: t\in\mathcal{B},\,i\in[D]\}$:
\begin{equation}
a(h_t)_i \;=\; \tilde z_{t,i}\cdot\mathbb{1}\!\bigl[(t,i)\in\mathrm{top}_{Bk}(\{\tilde z\})\bigr].
\end{equation}
Training regularization and per-policy widths are in Appendix~\ref{app:sae-details}.

\begin{figure*}[!t]
\centering
\includegraphics[width=\textwidth]{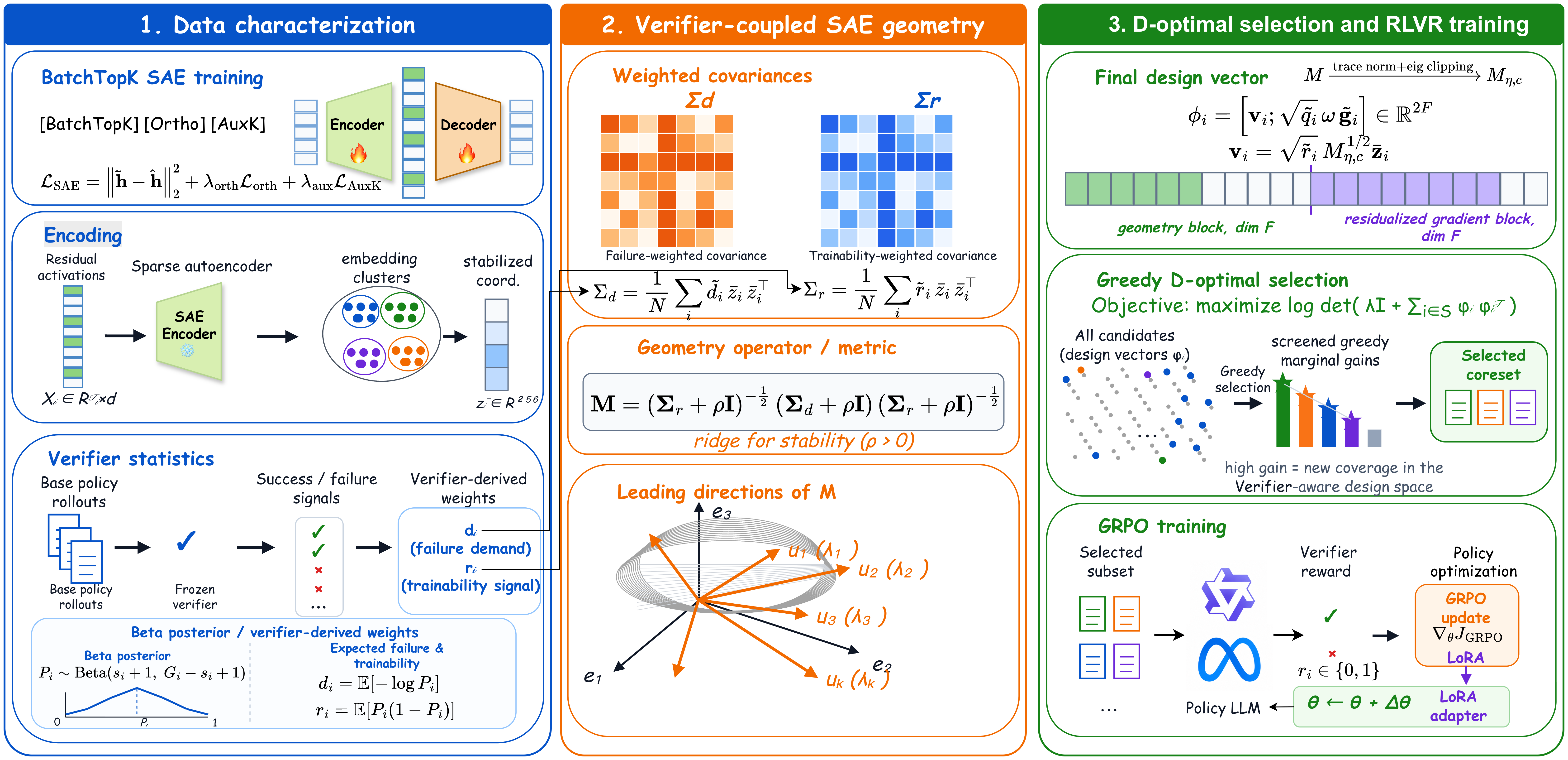}
\caption{\textbf{\method{} overview.}
\textbf{1. Data characterization:} train a BatchTopK SAE on frozen residual activations, encode problems into stabilized SAE-cluster coordinates, and derive verifier weights $d_i,r_i$ from base-policy rollouts.
\textbf{2. Verifier-coupled SAE geometry:} form the failure-weighted and trainability-weighted covariances $\Sigma_d,\Sigma_r$, combine them into the ridge-regularized metric $M$, and use its leading directions as the verifier-aware geometry.
\textbf{3. D-optimal selection and RLVR training:} apply trace normalization and eigenvalue clipping to obtain $M_{\eta,c}$, stack the geometry block with the residualized gradient block in $\phi_i$, greedily maximize the log-determinant objective, and train GRPO on the selected coreset.}
\label{fig:overview}
\end{figure*}

\section{Method}
\label{sec:method}

Given the pool $\mathcal{D}\!=\!\{x_i\}_{i=1}^N$, the verifier counts $\{s_i\}$ from \S\ref{sec:problem-setup}, and a target budget $K\!\ll\!N$, \method{} returns a size-$K$ subset $S^\star\!\subseteq\![N]$ following the pipeline in Figure~\ref{fig:overview}. We represent each instance on an SAE cluster basis (\S\ref{sec:sae-clusters}), construct a verifier-coupled coverage metric on that basis whose leading directions concentrate difficulty relative to trainability mass (\S\ref{sec:targetmetric}), optionally augment the per-instance design vector with a residualized gradient block (\S\ref{sec:gradient-block}), and select a non-redundant subset by greedy log-determinant maximization on the resulting design vectors (\S\ref{sec:subset-selection}). Numerical settings and full pseudocode are in the appendix.

\subsection{SAE Cluster Coordinates}
\label{sec:sae-clusters}

We want an instance coordinate that is more semantic than surface features such as length or LaTeX density, yet more inspectable than dense hidden states.Figure~\ref{fig:overview}(panel 1) illustrates the pipeline.  The sparse code introduced in Section~\ref{sec:problem-setup} provides one, since many of the dictionary directions correspond to recognizable concepts.

\paragraph{SAE training.}
We train one BatchTopK SAE per base model on that model's final-layer activations, harvested from OpenMathInstruct-2~\citep{toshniwal2024openmathinstruct} as an unsupervised activation source. The SAE optimizes only activation reconstruction and never sees verifier outcomes, evaluation labels, or \datasetname{DeepScaleR} selection statistics, so even if a prompt overlap exists between OpenMathInstruct-2 and the evaluation suite the dictionary cannot leak labels into selection. Throughout the paper we use a shared expansion factor of $32$ and average per-token sparsity $k{=}128$; per-policy widths, regularization details, and held-out reconstruction quality are in Appendix~\ref{app:sae-details}.

\paragraph{From sparse latents to cluster vectors.}
Individual SAE latents are too fine-grained for a corpus-level coordinate, since a single $40$K-instance pool typically activates several tens of thousands of distinct latents and any individual one fires on only a small fraction of instances. We therefore cluster related latents into $F{=}256$ semantic clusters before the selection stage. Each kept latent is embedded by a graph-hybrid representation that concatenates a \emph{presence direction} (from co-activation cosine similarity over instance-presence indicators) with a \emph{residual direction} (from residual mass variation after regressing out the presence signal); we then run spherical $K$-means on this embedding with $F{=}256$ clusters under a fixed seed ($F$ is distinct from both the selection budget $K$ of Eq.~\eqref{eq:dopt} and the BatchTopK sparsity $k{=}128$). Each instance $x_i$ is summarized by the per-cluster activation mass $m_i\in\R_{\ge 0}^{F}$ obtained as $m_{if} = \sum_{\ell\in\mathcal{C}_f} f_{i\ell}$ where $f_{i\ell} = \mathrm{mean}_t\,a(h_t)_\ell$ averages each latent's activation over the tokens of $x_i$ and $\mathcal{C}_f$ indexes the latents assigned to cluster $f$. Each cluster admits a short human-readable label, such as circle--coordinate geometry or polynomial-root algebra, which lets a researcher audit on which content directions the selected subset spends its budget. The full activation pipeline and clustering recipe are in Appendix~\ref{app:sae-details} and~\ref{app:cluster-construction}.

\subsection{Verifier-Coupled SAE Coverage}
\label{sec:targetmetric}

With the cluster coordinate fixed, we now turn it into a coverage objective for RLVR.As shown in Figure~\ref{fig:overview}(panel2), selection must satisfy two conditions on each picked instance that the cluster coordinate alone does not capture: the instance should be one the policy currently fails on, so training has something to repair, and its rollouts should produce a non-degenerate group-relative gradient, so the optimizer can act on the signal. Both conditions are functions of the per-instance success count $s_i$ rather than of the cluster coordinate $m_i$, and pull on different parts of the success-rate range, with the first condition pulling toward low $s_i$ and the second toward the mid-band where the gradient is largest. The construction below derives one weight per condition, combines them on the cluster coordinate $\bar z_i$, and produces a coverage metric on $\R^F$ whose leading directions concentrate failure beyond what already-trainable instances cover.

\paragraph{Difficulty weight and trainability weight.}
For each instance $x_i$ the verifier count $s_i$ induces a Beta posterior on its success rate, $P_i\sim\mathrm{Beta}(s_i{+}1,G_i{-}s_i{+}1)$, from which we derive two scalar weights with closed forms in the digamma function $\psi$ and the Beta moments:
\begin{equation}
\begin{aligned}
d_i &= \E[-\log P_i] = \psi(G_i{+}2)-\psi(s_i{+}1),\\
r_i &= \E[P_i(1{-}P_i)] = \frac{(s_i{+}1)(G_i{-}s_i{+}1)}{(G_i{+}2)(G_i{+}3)}.
\end{aligned}
\label{eq:dr}
\end{equation}
We call $d_i$ the \emph{difficulty weight}: it is monotonically decreasing in $s_i$, so it is high when the model fails often and low when it succeeds often. We call $r_i$ the \emph{trainability weight}: it is unimodal with peak near $s_i\!\approx\!G_i/2$, exactly where group-relative training signal is strongest. Concretely, the response-level advantage variance under group-relative optimization contains a $P_i(1{-}P_i)$ factor~\citep{zeng2025cures1037,gao2025prompt1135}, and our empirical per-instance gradient norms peak near intermediate success rates, so weighting by $r_i$ downweights both the all-fail tail (where the gradient vanishes) and the all-solve tail (where there is nothing to learn). We mean-normalize each weight across the corpus to keep them on a common scale and write the resulting weights $\tilde d_i,\tilde r_i$ below. Because $d_i$ and $r_i$ are independent functions of the same success rate, they let the method treat which content needs coverage separately from how trainable the rollouts are.

\paragraph{Verifier-coupled metric.}
We seek a metric on the SAE coordinate space that lifts directions where verifier failures concentrate, relative to what high-trainability instances already cover.
Let $\bar z_i\in\R^F$ denote a regularized version of the centered cluster mass $m_i$, detailed below.
On this space we form the weighted covariances
\begin{equation}
\Sigma_d=\frac1N\sum_i \tilde d_i \bar z_i\bar z_i^\top,\quad
\Sigma_r=\frac1N\sum_i \tilde r_i \bar z_i\bar z_i^\top,
\end{equation}
and the whitened-coverage metric
\begin{equation}
M=(\Sigma_r+\rho I)^{-1/2}(\Sigma_d+\rho I)(\Sigma_r+\rho I)^{-1/2}.
\label{eq:metric}
\end{equation}
$\Sigma_d$ and $\Sigma_r$ are SAE covariances reweighted by the difficulty and trainability weights respectively, so whitening the former by the latter isolates SAE directions where difficulty exceeds trainability rather than directions that are simply hard or frequent. The leading directions of $M$ maximize the regularized generalized Rayleigh quotient $v^\top(\Sigma_d+\rho I) v / v^\top(\Sigma_r+\rho I) v$, which is large on directions where difficulty mass exceeds trainability mass under the same ridge regularization. We then turn this metric into a per-instance design vector. In practice we apply finite-sample regularization to $M$ (trace normalization and bounded eigenvalue clipping with parameters $\eta,c$, denoted $M_{\eta,c}$ in Appendix~\ref{app:algorithm}); we also center $\bar z_i$ as described in the appendix. For notational simplicity, we write $M$ for this regularized version in the main text and define the design vector with the trainability weight:
\begin{equation}
v_i=\sqrt{\tilde r_i}\,M^{1/2}\bar z_i,
\label{eq:vi}
\end{equation}
so that instances contribute most to the $\log\det$ objective when they carry trainability mass and load on directions where difficulty most exceeds trainability. Regularization and ridge constants are listed in Appendix~\ref{app:metric-ablations} and Table~\ref{tab:method-hparams}.

\subsection{Gradient-Focused Block}
\label{sec:gradient-block}

The SAE coordinate captures content but is blind to update-direction information that the policy gradient would expose. To complement it, we add an optional auxiliary block whose contribution is concatenated to $v_i$ and turned off by setting its scalar mixing weight $\omega$ (Eq.~\eqref{eq:phi}) to zero.

For each instance $x_i$ we compute the per-instance negative log-likelihood (NLL) of the frozen base model on its own greedy continuation $y_i^{\mathrm{ref}}$ ($\mathcal{L}_i = -\sum_t \log\pi_\theta(y_{i,t}^{\mathrm{ref}}\mid x_i, y_{i,<t}^{\mathrm{ref}})$), backpropagate to the full transformer-block parameters, and form a per-instance gradient signature $g_i\in\R^{p_g}$ via TRAK-style per-block Johnson--Lindenstrauss random projection~\citep{park2023trak} ($p_g{=}L\!\cdot\!64$ where $L$ is the transformer-block count, so $p_g\!\in\!\{1792,2304,2048\}$ for the three models; see Appendix~\ref{app:gradient-pipeline}). The gradient block then applies two operations to this vector. First, we ridge-residualize against the SAE coordinate so the block contributes only directions not already spanned by $\bar z_i$:
\begin{equation}
\mathbf{G}^{\perp}=\mathbf{G}-\bar Z\bigl(\bar Z^\top\bar Z+\rho_g I\bigr)^{-1}\bar Z^\top \mathbf{G},
\label{eq:gresid}
\end{equation}
where $\mathbf{G}\in\R^{N\times p_g}$ and $\bar Z\in\R^{N\times F}$ stack the gradients and SAE coordinates row-wise (we use bold $\mathbf{G}$ for the gradient matrix to distinguish it from the rollout count $G$). After outlier clipping, PCA-whitening, and trace normalization to match the SAE block, we obtain $\widetilde g_i\in\R^{p_g}$. Second, we attach a per-instance weight $\tilde q_i \propto \|g_i^{\perp}\|^{-\alpha}$. Empirically on \datasetname{DeepScaleR}, large residual-gradient norms are dominated by short instances or saturated ones ($s_i\!=\!G_i$, all rollouts succeed), whereas focused reasoning instances occupy a lower-norm band; the inverse-norm factor therefore downweights this high-norm tail and redirects selection toward the lower-norm band. The specific per-instance norm ranges are reported in Appendix~\ref{app:gradient-pipeline}. Combining the two, the final design vector stacks the SAE block and the weighted gradient block:
\begin{equation}
\phi_i=\begin{bmatrix}\sqrt{\tilde r_i}\,M^{1/2}\bar z_i\\[2pt]\sqrt{\tilde q_i\,\omega}\,\widetilde g_i\end{bmatrix}\in\R^{F+p_g},
\label{eq:phi}
\end{equation}
in which the scalar $\omega$ trades off SAE-block coverage against gradient-block coverage and $\omega{=}0$ recovers an SAE-only design. The full $(\omega,\alpha)$ sweep on selection-time coverage diagnostics is in Appendix~\ref{app:omega-alpha-sweep}.

\subsection{Subset Selection}
\label{sec:subset-selection}

As shown in Figure~\ref{fig:overview}(panel 3), Given the design vector $\phi_i$, we want a subset whose rows span as many independent directions as possible, so that the budget is not wasted on instances that are mutually redundant. D-optimal design~\citep{kiefer1960equivalence,pukelsheim2006optimal} captures this preference through the log-determinant objective
\begin{equation}
S^\star=\arg\max_{|S|=K}\log\det\!\Bigl(\lambda I+\sum_{i\in S}\phi_i \phi_i^\top\Bigr),
\label{eq:dopt}
\end{equation}
which under a linear-Gaussian model is proportional to the expected information gain about the coverage direction~\citep{chaloner1995bayesian}. Let $A_S=\lambda I+\sum_{i\in S}\phi_i\phi_i^\top$ denote the running Gram matrix. Since the objective is monotone submodular, a greedy procedure that at each step adds the instance with the largest marginal gain $\Delta_i(S)=\log(1+\phi_i^\top A_S^{-1}\phi_i)$ attains the standard $(1{-}1/e)$ guarantee~\citep{nemhauser1978submodular,krause2008near}, and Sherman--Morrison rank-one updates on $A_S^{-1}$ keep the per-step cost at $O(p^2)$ rather than $O(p^3)$. Algorithm~\ref{alg:irds} summarizes the full pipeline; Appendix~\ref{app:diversity-baselines} compares against four classical alternatives on the same SAE coordinate ($k$-means clustering, facility-location coverage, statistical leverage scores, and a LESS-style~\citep{xia2024less} influence proxy).

\begin{algorithm}[t]
\caption{\method{}: verifier-coupled SAE subset selection. Full pseudocode in Appendix~\ref{app:algorithm}.}
\label{alg:irds}
\small
\begin{algorithmic}[1]
\Require Instances $\{x_i\}$; verifier counts $(s_i,G_i)$; cluster mass $\{m_i\}$; gradients $\{g_i\}$; budget $K$.
\State Posterior weights $\tilde d_i,\tilde r_i$ from $(s_i,G_i)$.
\State Regularize $m_i \to \bar z_i$; form $\Sigma_d,\Sigma_r$; whiten to $M$; set $v_i = \sqrt{\tilde r_i}\,M^{1/2}\bar z_i$.
\State Residualize $g_i^\perp = g_i - \bar Z(\bar Z^\top\bar Z+\rho_g I)^{-1}\bar Z^\top g_i$; trace-normalize to $\widetilde g_i$; concatenate $\phi_i = [v_i;\,\sqrt{\tilde q_i\,\omega}\,\widetilde g_i]$.
\State Greedily build $S^\star$ by the largest log-det marginal gains, using Sherman--Morrison rank-one updates.
\State \Return $S^\star$.
\end{algorithmic}
\end{algorithm}

\section{Experiments}
\label{sec:exp}

Our experiments answer five questions about \method{} under a shared GRPO recipe: (i) do its selected subsets translate into final-policy accuracy gains over scalar, diversity, and trajectory baselines (\S\ref{sec:main-results}); (ii) which design components in Section~\ref{sec:method} are responsible for those gains (\S\ref{sec:framework-axiom-empirical}); (iii) do the gains come at an out-of-domain cost (\S\ref{sec:ood-eval}); (iv) does the selected subset reflect content directions rather than surface artifacts (\S\ref{sec:case-study}); and (v) is the picture stable across data-budget choices (\S\ref{sec:budget-sweep}).

\subsection{Experimental Setup}
\label{sec:setup}

\paragraph{Models and training.}
We post-train three instruction-tuned models, \modelname{Qwen3-4B}, \modelname{Qwen3-1.7B}, and \modelname{Llama-3.1-8B-Instruct}, on the \datasetname{DeepScaleR} math-reasoning corpus~\citep{deepscaler2025} under a shared GRPO~\citep{shao2024deepseekmath} recipe with LoRA~\citep{hu2022lora} adapters and AdamW~\citep{loshchilov2019adamw} optimization. Each baseline picks its own $20\%$ training subset under this recipe so that only the data composition varies, and each model uses its own SAE and verifier-scored rollouts to instantiate the per-instance weights. Full hyperparameters and the SAE training recipe are in Appendix~\ref{app:rlvr-details} and~\ref{app:sae-details}.

\paragraph{Evaluation and baselines.}
We evaluate each trained policy on a six-benchmark math suite drawn from \datasetname{MATH500}, \datasetname{AIME24}, \datasetname{AIME25}, \datasetname{AMC23}, \datasetname{Minerva Math}, and \datasetname{OlympiadBench} (Appendix~\ref{app:datasets-eval}). For each problem we draw $n{=}16$ stochastic samples at temperature $0.6$, top-$p{=}0.95$, with a $4$k response budget, and score each sample as correct/incorrect against the ground-truth answer; mean@$16$ reports the average per-problem pass rate over the $16$ samples, then averaged across the problem set. We compare against \baseline{Random} sampling, perplexity-pruned subsets (\baseline{PPL-top}, \baseline{PPL-middle},~\citealp{ankner2024perplexed}), instruction-following difficulty (\baseline{IFD},~\citealp{li2023ifd}), token-length ranking (\baseline{Token-length},~\citealp{xia2024rethinking}), trajectory-based \baseline{LIMR}~\citep{li2025limr}, and the offline density/PageRank/determinantal-point-process (DPP) method \baseline{DEPO}~\citep{tang2025towards1321}.

\subsection{Main Results}
\label{sec:main-results}

\method{} reaches the highest Overall accuracy on each of the three models (Table~\ref{tab:main-suite-by-model-selection}), scoring $67.0$ on \modelname{Qwen3-4B}, $54.2$ on \modelname{Qwen3-1.7B}, and $22.6$ on \modelname{Llama-3.1-8B-Instruct}, with Overall improvements of $+3.9$, $+4.0$, and $+0.5$\,pp over the strongest baseline. On the larger \datasetname{MATH500} subset (500 problems), gains are $+2.2$\,pp on \modelname{Qwen3-4B} ($91.3$ vs.\ \baseline{LIMR} $89.1$) and $+3.5$\,pp on \modelname{Qwen3-1.7B}. The largest per-benchmark gains occur on the competition subsets, which we note are also the smallest test sets (\datasetname{AIME24}: $60$, \datasetname{AIME25}: $30$, \datasetname{AMC23}: $46$ problems): on \modelname{Qwen3-4B}, \method{} gains $+11.4$\,pp on \datasetname{AIME24} ($42.0$ vs.\ \baseline{LIMR} $30.6$), $+7.1$\,pp on \datasetname{AMC23} ($79.5$ vs.\ \baseline{PPL-top} $72.4$), and $+6.6$\,pp on \datasetname{AIME25} ($35.6$ vs.\ \baseline{DEPO} $29.0$); \modelname{Qwen3-1.7B} matches this pattern on the same three subsets ($+6.1$/$+6.9$/$+7.1$\,pp). On \modelname{Llama-3.1-8B-Instruct} the Overall margin is smaller ($+0.5$\,pp) and individual baselines lead on the smaller competition subsets.

\input{tables/main_suite_by_model_selection}

\subsection{Ablations}
\label{sec:framework-axiom-empirical}

Components contribute unequally but each is non-redundant. To attribute the headline gain to specific design choices, we train six component-removal variants under the shared recipe (Table~\ref{tab:ablation-compact}; per-subset breakdown in Appendix~\ref{app:ablation-detail}). The ablation reveals an asymmetric structure across the three models. The trainability weight $r_i$ is the \emph{structurally critical} component: replacing the pair by $d_i$ alone (i.e.\ removing $r_i$) collapses the method by $4.1$ to $19.1$\,pp because pure difficulty-mass selection concentrates the budget on all-fail instances where group-relative advantage variance is near zero. The difficulty weight $d_i$ is a \emph{smaller but non-trivial refinement}: replacing the pair by $r_i$ alone (i.e.\ removing $d_i$) costs $0.7$ to $3.2$\,pp because $r_i$ already selects the mid-success band that gradient-based training favors, but lacks $d_i$'s drive toward the SAE directions where the policy still fails most. The whitened metric $M$ and the set-level greedy procedure each carry the next-largest fraction of the lift: setting $M{=}I$ drops Overall by up to $-9.2$\,pp, and substituting the pointwise product $d_i r_i$ for the set-level objective costs up to $-6.6$\,pp, so both whitening and the redundancy penalty matter. The SAE cluster basis is itself load-bearing on \modelname{Qwen3-4B}, where replacing it with a PCA-$256$ projection of dense gradients at the same dimension loses $-4.5$\,pp Overall and $-10.0$\,pp on \datasetname{AIME24}; on the other two models the same substitution costs $1.1$ to $2.1$\,pp, indicating that SAE content structure is most discriminative on the strongest base policy. The gradient block contributes the smallest margin ($0.7$ to $2.8$\,pp); the SAE-only variant ($\omega{=}0$) already attains $64.2/53.0/21.9$ Overall on the three models, so the SAE-coupled coverage objective carries the bulk of the lift and the gradient block is an optional refinement. Across the three budgets reported in \S\ref{sec:budget-sweep}, the structural rank order ($r_i$ removal $\gg$ $M{=}I$ $\gg$ $d_i r_i$ pointwise $\gg$ $d_i$/gradient/PCA-$256$) holds at every budget on every model, so these deltas reflect stable component contributions rather than budget-specific noise. Per-subset breakdowns are in Appendix~\ref{app:ablation-detail}; the $(\omega,\alpha)$ sweep, reported on selection-time coverage diagnostics rather than downstream accuracy, is in Appendix~\ref{app:omega-alpha-sweep}.
\input{tables/ablation_compact}

\vspace{-0.4em}
\subsection{Out-of-Domain Transfer}
\label{sec:ood-eval}
\vspace{-0.2em}
We evaluate every trained checkpoint on HumanEval~\citep{chen2021humaneval}, LiveCodeBench~\citep{jain2025livecodebench}, and ReClor~\citep{yu2020reclor} to span code, contest coding, and reading-comprehension reasoning (Table~\ref{tab:ood-compact}; per-task in Appendix~\ref{app:ood-transfer-detail}). \method{} attains the best three-task average on \modelname{Llama-3.1-8B-Instruct} ($48.2$ vs.\ $47.9$) and \modelname{Qwen3-1.7B} ($49.9$ vs.\ $49.5$), and is $1.0$\,pp below \baseline{Random} on \modelname{Qwen3-4B} ($64.9$ vs.\ $65.9$, concentrated on \datasetname{LiveCodeBench}), about a fourth of the corresponding $+3.9$\,pp in-domain lift.
\input{tables/ood_compact}

\subsection{Auditing the Selected Subset}
\label{sec:case-study}

We check that the selected subset reflects content directions rather than surface features via four diagnostics that all agree across the three models: high-difficulty localization, cross-policy consistency, surface-artifact non-explanation, and label-shuffle falsification.

\paragraph{High-difficulty localization.}
The leading eigenvectors of $M$ load on instances with unusually low verifier success: on each of the three models, the median success count of the top-$25$ highest-projection instances is well below the corpus median, so the directions on which the method spends its budget coincide with the high-difficulty regions the metric is meant to lift.

\paragraph{Cross-policy consistency.}
On each of the three policies' independently trained SAE bases, the leading axes of $M$ label the same set of recognizable problem motifs (geometry, divisor counting, polynomial roots, multiple-choice stems), so the method reaches comparable content directions across architectures.

\paragraph{Surface-artifact non-explanation.}
A linear regression of the selection indicator on six surface features (length, bracket count, digit density, multiple-choice stem indicator, and two notation-density statistics) explains a negligible fraction of the indicator variance on every model. Conversely, ablating the verifier-coupled SAE block (i.e.\ replacing $M^{1/2}\bar z_i$ with $\bar z_i$) raises the per-instance marginal-gain variance several-fold above a mass-matched random control. Together these two checks are inconsistent with a purely surface-driven explanation of the lift.

\paragraph{Label-shuffle falsification.}
A global verifier-label shuffle collapses the top eigenvalue of $M$ and drives the leading-subspace overlap with the true-label $M$ to near zero, which is the falsification check we expect to pass and which the method indeed passes.

Concrete examples in Appendix~\ref{app:qualitative-audits} illustrate the method-level mechanism: \method{} picks a high-difficulty multiple-choice question (MCQ) stem exemplar but rejects a near-duplicate from the same template, and routes a contrastive number-theory cluster to a mid-success instance with non-zero trainability weight. A full-spectrum feature-amplification scan (residual-stream injection along each of the $256$ $M$-eigenvectors at varying scales) and a base-vs-trained spectrum-invariance result are in Appendix~\ref{app:spectrum-interp}.

\subsection{Budget Sensitivity}
\label{sec:budget-sweep}

\method{}'s lift is robust across data budgets. Training every (method, model) pair at $10\%$ and $30\%$ of \datasetname{DeepScaleR} under the same recipe (Table~\ref{tab:main-suite-budget-compact}), \method{} retains the highest Overall accuracy at every budget on the three models. Per-benchmark numbers at $10\%$ and $30\%$ are in Appendix~\ref{app:budget-sweep-detail}.
\input{tables/main_suite_budget_compact}

\section{Conclusion}

We presented \method{}, an offline RLVR data-selection method that picks a non-redundant subset on a sparse-autoencoder cluster basis through a verifier-coupled coverage objective. The metric $M$ disentangles where the model still fails ($d_i$) from where it can still learn ($r_i$), and greedy log-determinant maximization turns these per-instance signals into a set-level selection that pays for each weakness only once. Across three instruction-tuned models and six math benchmarks, \method{} attains the highest Overall accuracy at an order-of-magnitude lower selection cost than the trajectory-based baseline, with a small in-domain margin on \modelname{Llama-3.1-8B-Instruct} and a small out-of-domain regression on \modelname{Qwen3-4B}. By selecting on a human-readable SAE basis, \method{} shows that the verifier-coupled component of RLVR data selection can be made auditable without losing in-domain accuracy.

\section*{Limitations}

\method{} estimates the verifier-coupled weights once under the frozen model, so the failure profile that the method targets can drift away from the model's actual profile after many updates; a semi-online refresh of these weights on top of the fixed SAE cluster basis is a natural extension that we leave to future work. The method also requires a model-specific SAE trained on the model's final-layer activations, which restricts its application to settings where such activations are accessible. We validate the method only on math RLVR under GRPO, so transfer to non-math RLVR domains and to non-GRPO algorithms remains to be demonstrated. Finally, all reported numbers come from a single training seed per cell, so seed-level RL variance is not captured.

\section*{Ethics Statement}

\method{} selects existing instances from publicly released math datasets (\datasetname{DeepScaleR}~\citep{deepscaler2025}, OpenMathInstruct-2~\citep{toshniwal2024openmathinstruct}) and does not generate, label, or release new problems. All models and benchmarks listed in \S\ref{sec:setup} are public and used under their original licenses; no human subjects or personally identifying data are involved. The method amplifies instances on which a model already fails, so its intended use is improving the math-reasoning accuracy of post-trained models; \method{} does not change the verifier, reward signal, or policy architecture, and we do not foresee dual-use risks beyond those general to RLVR post-training.

\bibliography{custom}

\clearpage
\appendix

\section{Component-removal ablation: per-subset detail}
\label{app:ablation-detail}

Table~\ref{tab:ablation-3policy} expands the body Table~\ref{tab:ablation-compact} with the per-subset mean@$16$ on all six benchmarks for every component-removal variant and every model at the shared $20\%$ budget.

\input{tables/ablation_3policy_projected}

\section{Metric rationale and greedy selection guarantee}
\label{app:theory}

This appendix consolidates the design rationale for Section~\ref{sec:method}: the spectral interpretation of the verifier-coupled metric $M$ together with the per-instance design vector $v_i$, and the $(1{-}1/e)$ submodular guarantee for the greedy procedure.

\subsection{Spectral decomposition of $M$ and $v_i$}
\label{app:theory-rayleigh}

\begin{lemma}[Generalized Rayleigh form of $M$]
Let $\widetilde\Sigma_r=\Sigma_r+\rho I$ and $\widetilde\Sigma_d=\Sigma_d+\rho I$ be the ridge-regularized covariances from Section~\ref{sec:targetmetric}, and let $M=\widetilde\Sigma_r^{-1/2}\widetilde\Sigma_d\widetilde\Sigma_r^{-1/2}$. The eigenpairs $(\lambda_k,u_k)$ of $M$ are in one-to-one correspondence with the solutions of the generalized eigenvalue problem $\widetilde\Sigma_d w_k=\lambda_k\widetilde\Sigma_r w_k$ via $w_k=\widetilde\Sigma_r^{-1/2}u_k$, and the leading direction maximizes the regularized Rayleigh quotient
\begin{equation}
\rho(w)=\frac{w^\top\widetilde\Sigma_d w}{w^\top\widetilde\Sigma_r w}.
\label{eq:rayleigh}
\end{equation}
\end{lemma}

The lemma identifies $M$'s top eigenvectors as the SAE coordinate directions where difficulty mass most exceeds trainability mass: directions with comparable difficulty-weighted and trainability-weighted covariance energy receive quotient near one, while directions with excess difficulty relative to trainability mass receive larger eigenvalues. Writing $M=\sum_k \lambda_k u_k u_k^\top$, the per-instance design vector $v_i=\sqrt{\tilde r_i}\,M^{1/2}\bar z_i$ from Eq.~\eqref{eq:vi} then satisfies
\begin{equation}
\|v_i\|^2 = \tilde r_i \sum_k \lambda_k\,(\bar z_i^\top u_k)^2,
\label{eq:vi-decomp}
\end{equation}
so an instance contributes to the objective in proportion to (i) its trainability weight $\tilde r_i$ and (ii) the energy it places on the leading $u_k$ weighted by $\lambda_k$. Instances that are either untrainable ($\tilde r_i\!\to\!0$) or that load primarily on low-$\lambda_k$ axes have limited design energy regardless of their total cluster mass.

\subsection{Greedy D-optimal guarantee}
\label{app:theory-greedy}

Let $d_\phi=\dim(\phi_i)$ ($d_\phi\!=\!F$ in the SAE-only regime and $d_\phi\!=\!F+p_g$ when the gradient block is active), and define the set function
\begin{equation}
J(S)=\log\det\!\Bigl(\lambda I_{d_\phi}+\sum_{i\in S}\phi_i\phi_i^\top\Bigr).
\label{eq:set-fn}
\end{equation}
Since $\lambda I_{d_\phi}\succ 0$, the normalized objective $\bar J(S)=J(S)-J(\emptyset)$ with $J(\emptyset)=d_\phi\log\lambda$ is monotone submodular~\citep{nemhauser1978submodular,krause2008near}, so the greedy procedure of Section~\ref{sec:subset-selection} that adds at each step the instance with the largest marginal gain $\Delta_i(S)=\log\bigl(1+\phi_i^\top A_S^{-1}\phi_i\bigr)$ attains
\begin{equation}
J(S_{\mathrm{greedy}})-J(\emptyset)\geq (1-1/e)\bigl(J(S^\star)-J(\emptyset)\bigr)
\label{eq:greedy-bound}
\end{equation}
relative to the optimal size-$K$ subset $S^\star$. The implementation in Algorithm~\ref{alg:irds-full} uses a screened candidate queue of size $Q{=}1024$ refreshed every $R{=}256$ additions, which matches the exact greedy objective within $0.1\%$ on held-out checks.

\section{Out-of-domain transfer: per-task detail}
\label{app:ood-transfer-detail}

Table~\ref{tab:ood-eval} expands the body Table~\ref{tab:ood-compact} with per-task results on HumanEval (HE)~\citep{chen2021humaneval} for Python code generation, LiveCodeBench (LCB)~\citep{jain2025livecodebench} for contest-level coding, and ReClor (RC)~\citep{yu2020reclor} for reading-comprehension multiple choice. Decoding settings match the in-domain protocol (mean@$16$ for HE/RC, pass@$1$ for LCB) under each benchmark's standard prompt format; no system prompt is injected.

\input{tables/ood_eval}

\section{Budget-sweep per-benchmark detail}
\label{app:budget-sweep-detail}

The body Table~\ref{tab:main-suite-budget-compact} reports Overall mean@$16$ at the three measured budgets ($10/20/30\%$). Tables~\ref{tab:main-suite-at-10pct} and~\ref{tab:main-suite-at-30pct} below give the full per-benchmark mean@$16$ breakdown at $10\%$ and $30\%$ budgets, with all seven baseline methods and \method{} on all three models and all six benchmarks.

\input{tables/main_suite_at_10pct}
\input{tables/main_suite_at_30pct}

\section{Implementation Details}
\label{app:implementation}

\subsection{Algorithm implementation}
\label{app:algorithm}

Algorithm~\ref{alg:irds-full} gives the full pseudocode behind the body skeleton in Algorithm~\ref{alg:irds}, taking the encoded cluster masses $\{m_i\}$ (Appendix~\ref{app:sae-details}) and per-instance gradients $\{g_i\}$ (Appendix~\ref{app:gradient-pipeline}) as inputs. The greedy D-opt solver uses $\lambda{=}1$, a screened candidate queue of size $Q{=}1024$ (refreshed every $R{=}256$ additions), and rank-one Sherman--Morrison updates of $A^{-1}$ to avoid recomputing the full inverse at each iteration. The full selection completes in under $60$ seconds on a single GPU.

\begin{algorithm*}[t]
\caption{\method{}: full pseudocode.}
\label{alg:irds-full}
\small
\begin{algorithmic}[1]
\Require Instances $\{x_i\}_{i=1}^{N}$ with verifier counts $(s_i, G_i)$, raw cluster masses $\{m_i\}\in\R^F$, per-instance TRAK NLL-gradient signatures $\{g_i\}\in\R^{p_g}$ (Appx.~\ref{app:gradient-pipeline}), budget $K$.
\Require Defaults $\rho{=}10^{-1}$, $\eta{=}0.5$, $c{=}2$, $\rho_g{=}10^{-3}$, $\omega{=}1$, $\alpha{=}1$, $\lambda{=}1$, $\varepsilon{=}10^{-3}$, screen size $Q{=}1024$, refresh $R{=}256$.
\Ensure Subset $S^\star\subseteq[N]$, $|S^\star|=K$.
\vspace{2pt}
\Statex \emph{Per-instance posterior weights.}
\State $d_i \gets \psi(G_i+2) - \psi(s_i+1)$
\State $r_i \gets (s_i+1)(G_i-s_i+1)/[(G_i+2)(G_i+3)]$
\State $\tilde d_i, \tilde r_i \gets$ mean-one rescaling
\Statex \emph{SAE metric block.}
\State Regularize $m_i$ to $\bar z_i$ (success-axis removal, $q_{99}$ row clip, bucket-mean subtraction)
\State $\Sigma_d \gets \tfrac{1}{N}\sum_i \tilde d_i\,\bar z_i\bar z_i^\top$,\quad $\Sigma_r \gets \tfrac{1}{N}\sum_i \tilde r_i\,\bar z_i\bar z_i^\top$
\State $M \gets (\Sigma_r+\rho I)^{-1/2}(\Sigma_d+\rho I)(\Sigma_r+\rho I)^{-1/2}$
\State $M_{\eta,c} \gets U\,\diag(\mathrm{clip}(\Lambda^\eta, 1/c, c))\,U^\top$ with $M = U\Lambda U^\top$, trace-normalized
\State $v_i \gets \sqrt{\tilde r_i}\, M_{\eta,c}^{1/2}\, \bar z_i$
\Statex \emph{Optional gradient block.}
\If{$\omega > 0$}
  \State $g_i^\perp \gets g_i - \bar Z(\bar Z^\top \bar Z + \rho_g I)^{-1} \bar Z^\top g_i$
  \State $\widetilde g_i \gets$ row-clip at $q_{99}$, unit-normalize, whitened PCA (ridge $10^{-4}$, drop ridge-floor eigendirections), trace-match to SAE block
  \State $\bar g \gets \tfrac{1}{N}\sum_j \|g_j^\perp\|$,\quad $\tilde q_i \gets (\bar g/\max(\|g_i^\perp\|, \varepsilon\bar g))^{\alpha}$
  \State $\phi_i \gets [\, v_i ;\, \sqrt{\tilde q_i\,\omega}\,\widetilde g_i\,] \in \R^{F+p_g}$
\Else
  \State $\phi_i \gets v_i \in \R^F$
\EndIf
\vspace{2pt}
\Statex \emph{Greedy D-optimal selection.}
\State $S \gets \emptyset$,\quad $A^{-1} \gets \lambda^{-1} I$
\For{$k = 1, \dots, K$}
  \State $\mathcal{Q}_k \gets $ top-$Q$ candidates by $\phi_i^\top A^{-1}\phi_i$ (refresh every $R$ steps)
  \State $i^\star \gets \arg\max_{i \in \mathcal{Q}_k \setminus S}\, \log(1 + \phi_i^\top A^{-1}\phi_i)$
  \State $S \gets S \cup \{i^\star\}$
  \State $A^{-1} \gets A^{-1} - \dfrac{(A^{-1}\phi_{i^\star})(A^{-1}\phi_{i^\star})^\top}{1 + \phi_{i^\star}^\top A^{-1}\phi_{i^\star}}$ \hfill (Sherman--Morrison)
\EndFor
\State \Return $S^\star \gets S$
\end{algorithmic}
\end{algorithm*}

\subsection{Method hyperparameters}
\label{app:method-hparams-app}

Table~\ref{tab:method-hparams} compiles the numerical settings of every method stage and the shared GRPO recipe; they are referenced individually in the corresponding subsections of Section~\ref{sec:method} and Appendix~\ref{app:rlvr-details}.
\input{tables/selector_hparams_compact}

\subsection{Per-instance gradient extraction}
\label{app:gradient-pipeline}

The gradient block in Eq.~\eqref{eq:phi} uses a single per-instance residualized gradient $g_i^\perp$ produced by the following one-pass pipeline against the frozen model, applied identically to \modelname{Qwen3-4B}, \modelname{Qwen3-1.7B}, and \modelname{Llama-3.1-8B-Instruct}.

\paragraph{Per-instance NLL gradient (TRAK signature).}
For each instance $x_i$, we sample one greedy continuation $y_i^{\mathrm{ref}}$ from the same frozen base model used by the verifier scoring step, and compute the per-token negative log-likelihood loss $\mathcal{L}_i=-\sum_{t}\log\pi_\theta(y_{i,t}^{\mathrm{ref}}\mid x_i,y_{i,<t}^{\mathrm{ref}})$. We backpropagate $\mathcal{L}_i$ once and read $\nabla_\theta \mathcal{L}_i$ off every transformer-block parameter (token embeddings, $\mathtt{lm\_head}$, and the final norm are excluded as they are dominated by token-identity effects and uninformative for problem-level coverage). For each transformer block $\ell\!\in\!\{1,\dots,L\}$ we concatenate the block's parameter gradients in a fixed name order into a single vector and Johnson--Lindenstrauss--project it to $p_\ell{=}64$ dimensions using a Gaussian random matrix with a layer-deterministic seed; the matrix is regenerated on the fly with bounded memory and shared across all instances, so projected signatures are directly comparable across $x_i$. This is the per-block TRAK projection of \citet{park2023trak}. Stacking across blocks gives the per-instance gradient signature $g_i\in\R^{p_g}$ with $p_g{=}L\!\cdot\!p_\ell$: $p_g{=}1792$ on \modelname{Qwen3-1.7B} ($L{=}28$), $p_g{=}2304$ on \modelname{Qwen3-4B} ($L{=}36$), and $p_g{=}2048$ on \modelname{Llama-3.1-8B-Instruct} ($L{=}32$). SAE coupling enters in the next step (residualization), not in the gradient signature itself.

\paragraph{Residualization, row-clip, and whitening.}
We then apply the four-step gradient block construction of Algorithm~\ref{alg:irds}: ridge residualization $g_i^\perp=g_i-\bar Z(\bar Z^\top\bar Z+\rho_g I)^{-1}\bar Z^\top g_i$ with $\rho_g{=}10^{-3}$, which removes from $g_i$ any direction linearly explained by the SAE coordinate $\bar z_i$; row-clip at the $99$th percentile of $\|g_j^\perp\|$ followed by row-renormalization to unit norm; PCA-whitening of the residual matrix (ridge $10^{-4}$, dropping eigendirections below the ridge floor); and trace-normalization so that $\tfrac1N\sum_i\|\widetilde g_i\|^2$ matches the SAE block at $\omega{=}1$. The mean-normalized inverse-norm weight $\tilde q_i = (\bar g/\max(\|g_i^\perp\|,\varepsilon\bar g))^\alpha$ with $\bar g = \tfrac{1}{N}\sum_j\|g_j^\perp\|$ and $\varepsilon{=}10^{-3}$ (and $\alpha{=}1$ in the headline runs), used in Eq.~\eqref{eq:phi}, is then applied as a scalar row weight on the rescaled $\widetilde g_i$. As a sanity check on the residualization, regressing $g_i$ on $\bar z_i$ across the 40K-instance pool yields $R^2{=}0.024$ globally on \modelname{Qwen3-1.7B}, with mean residual-norm reduction below $1\%$ across success-count buckets, so the gradient block enters $\phi_i$ as substantially independent information rather than a near-duplicate of the SAE coordinate.

\paragraph{High-gradient-norm tail.}
The residual-gradient norm $\|g_i^\perp\|$ on \datasetname{DeepScaleR} grows monotonically with the verifier success count $s_i$: on \modelname{Qwen3-1.7B} the per-bucket mean increases from $\|g_i^\perp\|{\approx}7.4$ at the all-fail tail ($s_i{=}0$) to $\|g_i^\perp\|{\approx}11.1$ at the saturated tail ($s_i{=}G$), with the saturated-tail $95$th percentile reaching ${\approx}21$ versus ${\approx}14$ at the mid-success bucket. The same monotone pattern holds on the other two models. The inverse-norm row weight $\tilde q_i$ in Section~\ref{sec:gradient-block} downweights this saturated tail---short, easy, and all-success instances dominate the high-norm region, so suppressing it routes the budget toward focused-reasoning instances at intermediate success rates.

\subsection{SAE training}
\label{app:sae-details}

Each model uses its own SAE trained on that policy's final-layer math activations; no SAE is shared across the three policies. We use shared expansion factor $32$ and sparsity $k{=}128$ throughout, yielding per-policy SAE widths $D{=}81{,}920$ / $65{,}536$ / $131{,}072$ for \modelname{Qwen3-4B} / \modelname{Qwen3-1.7B} / \modelname{Llama-3.1-8B-Instruct} ($d_{\mathrm{model}}{=}2{,}560$ / $2{,}048$ / $4{,}096$). The activation corpus is the OpenMathInstruct-2~\citep{toshniwal2024openmathinstruct} math-instruction collection ($300{,}000$ problems spanning grade-school to competition-level math, drawn from public sources curated by NVIDIA for math-reasoning post-training), used here purely as an activation source for unsupervised SAE training. We run each policy on these instances and harvest last-layer residual-stream token activations, sampling at most $512$ tokens per instance with seeded sampling for determinism; the resulting per-policy training set contains ${\sim}20{\rm M}$ tokens, written to disk in eight shards across $8\!\times$\,GPU. We remove known overlaps with the math evaluation suite when identifiable; the SAE receives no verifier reward, validation outcome, evaluation label, or downstream benchmark label, is fixed before selection, and is not updated during GRPO.

\begin{figure*}[t]
\centering
\includegraphics[width=\textwidth]{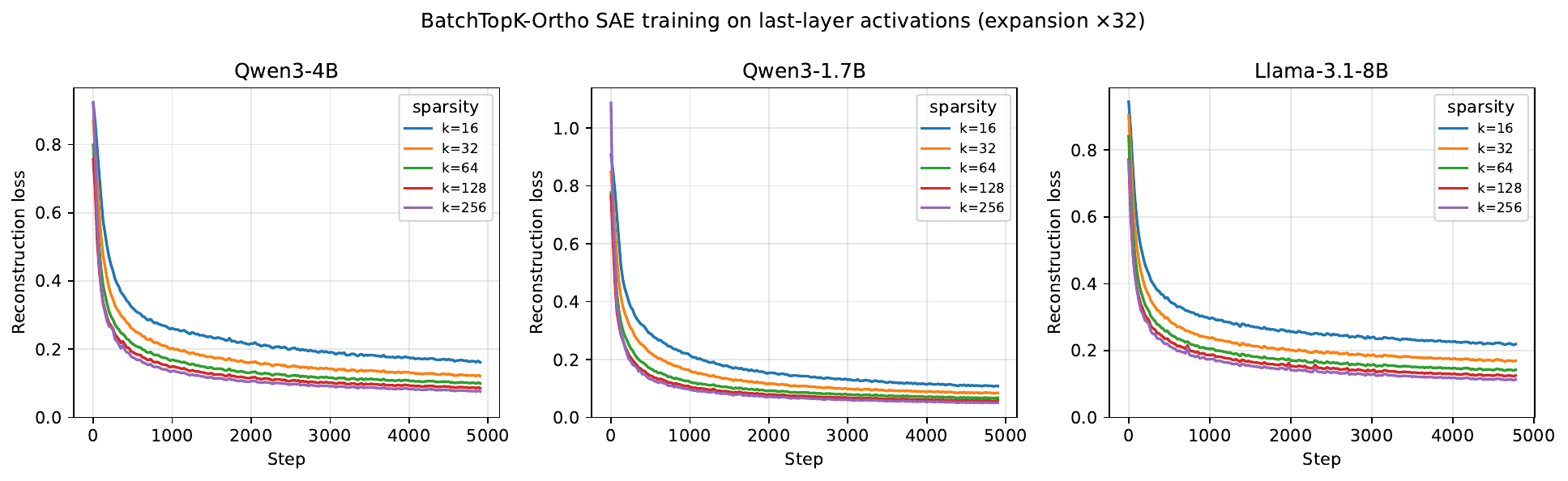}
\caption{BatchTopK-Ortho SAE training: held-out reconstruction loss vs.\ step, per model and per sparsity $k$.}
\label{fig:sae-training}
\end{figure*}

SAE training adds a decoder-orthogonality penalty $\lambda_{\mathrm{ortho}}\sum_{i\neq j}(W_{\mathrm{dec}}[:,i]^\top W_{\mathrm{dec}}[:,j])^2$ to discourage duplicate dictionary directions and an AuxK auxiliary loss~\citep{gao2024scalingsaes} to recover dead latents. Figure~\ref{fig:sae-training} reports the BatchTopK-Ortho sparsity sweep across all three model-specific SAEs at $k\!\in\!\{16,32,64,128,256\}$: held-out reconstruction loss decreases monotonically with $k$ on every model, and the orthogonality penalty keeps the duplicate / dead-feature rate low without harming convergence. We use $k{=}128$ as the shared setting. The selected \modelname{Qwen3-4B} SAE checkpoint is evaluated on $200$k held-out activation samples: normalized MSE $0.0834$, mean $L_0{=}128$, active-latent count $80{,}251$ of $81{,}920$, dead-feature rate $2.0\%$. End-to-end SAE training wall-clock on one $8{\times}$\,GPU node is $18$\,min for \modelname{Qwen3-1.7B}, $27$\,min for \modelname{Qwen3-4B}, and $72$\,min for \modelname{Llama-3.1-8B-Instruct}. We then encode the entire training set in eight shards; each instance has a nonnegative cluster-mass vector, used for selection after centering, residualization, and whitening and for interpretation in raw form.

\subsection{RLVR training and evaluation}
\label{app:rlvr-details}

We adopt GRPO~\citep{shao2024deepseekmath}, implemented in the VeRL framework~\citep{sheng2025verl}, as the default RL algorithm. \method{} is an instance-selection method that operates upstream of the RL loop and is algorithmically compatible with other on-policy RLVR algorithms such as DAPO~\citep{yu2025dapo}.
At each step we sample $G{=}8$ rollouts per instance at temperature $1.0$ and top-$p{=}1.0$, score with a rule-based binary verifier ($1$ if the response answer matches the ground-truth label, $0$ otherwise), and form group-relative advantages as in standard GRPO. Training runs for $2$ epochs on the selected $20\%$ subset, with instance batch size $128$, maximum input length $1{,}024$ tokens, and maximum response length $4{,}096$ tokens; all updates use LoRA adapters~\citep{hu2022lora} (rank $64$, scaling $\alpha_{\mathrm{LoRA}}{=}128$; distinct from the inverse-norm exponent $\alpha$ of Section~\ref{sec:gradient-block}) in bf16. Optimization is performed by AdamW~\citep{loshchilov2019adamw} ($\beta_1{=}0.9$, $\beta_2{=}0.999$, weight decay $0.1$) with learning rate $5\!\times\!10^{-6}$, $5$ warmup steps, and gradient clip norm $1.0$. The PPO clip ratio is $0.2$, the KL coefficient against the frozen reference policy is $10^{-3}$, and the entropy coefficient is $10^{-3}$. All experiments are conducted on $8\!\times$ NVIDIA H20 GPUs.

We monitor validation on \datasetname{MATH500} with $n{=}16$ samples per problem during training. Final evaluation uses identical decoding across all methods on the $3{,}230$-instance union of math benchmarks (Appendix~\ref{app:datasets-eval}), with $n{=}16$ samples per problem at temperature $0.6$, top-$p{=}0.95$, and a $4$k response budget; only the selected instance subset varies across methods.

\subsection{Chat template}
\label{app:chat-template}

Each problem is wrapped as a single user turn and tokenized through the model's native chat template (\modelname{Qwen3-4B}, \modelname{Qwen3-1.7B}, and \modelname{Llama-3.1-8B-Instruct} each use their own instruction template as released). We prepend a single system prompt that asks the model to produce a chain-of-thought followed by a boxed final answer; the rule-based verifier extracts the boxed expression and checks equivalence with the ground truth.

\medskip
\noindent\fbox{\parbox{0.98\columnwidth}{%
\small\textbf{System prompt (training and evaluation).}\\[2pt]
Please reason step by step, and put your final answer within \texttt{\textbackslash boxed\{\}}.
}}
\medskip

\noindent\fbox{\parbox{0.98\columnwidth}{%
\small\textbf{User turn.}\\[2pt]
\{problem statement from \datasetname{DeepScaleR}\}
}}
\medskip

The same template is used for selection-time verifier rollouts ($G{=}8$ per instance), for GRPO training rollouts, and for final benchmark evaluation, so the verifier always sees responses in the same format. Out-of-domain evaluation on HumanEval, LiveCodeBench, and ReClor uses each benchmark's standard prompt format (Section~\ref{sec:ood-eval}); no system prompt is injected on those tasks.

\input{analysis/family_construction_recipe}

\input{analysis/datasets_and_benchmarks}

\section{Baseline Methods}
\label{app:baseline-details}

All main baselines select the same $20\%$ budget from the same \datasetname{DeepScaleR} training set and are trained with the GRPO recipe in Appendix~\ref{app:rlvr-details}.
For response-dependent SFT-style methods, we use the provided answer or a single generated response, while offline RLVR methods use the same verifier-sampling records as \method{}.

\paragraph{Scalar and SFT-style methods.}
\baseline{Random} samples uniformly without replacement.
\baseline{Token-length} ranks by the total tokenized length of the problem and response, following the length-control use in SFT-scale data selection of \citet{xia2024rethinking}.
\baseline{PPL-top} and \baseline{PPL-middle} both follow the small-reference-model perplexity-pruning recipe of \citet{ankner2024perplexed}: a fixed scorer model assigns each instance a conditioned perplexity (mean negative log-likelihood of the response tokens given the instruction), and the method takes the top $20\%$ for \baseline{PPL-top} or the middle $20\%$ closest to the corpus median for \baseline{PPL-middle}.
\baseline{IFD} uses Instruction-Following Difficulty~\citep{li2023ifd}, computed as the discrepancy between the response likelihood with and without the instruction context.
For the \datasetname{DeepScaleR} training set, these scores use the provided problem-answer pairs as the response text.

\paragraph{RLVR and uncertainty methods.}
\baseline{LIMR} follows the learning-impact trajectory criterion of \citet{li2025limr}; all instances are scored after the same warmup and selected at the same $20\%$ budget.
\baseline{DEPO}~\citep{tang2025towards1321} is the Data-Efficient Policy Optimization pipeline of Tang et al., here labeled DEPO; it differs from the unrelated Difficulty-Estimated Policy Optimization of \citet{zhao2026difficulty6375} (which uses an online difficulty estimator) and which we discuss separately in Related Work. Tang-DEPO has both offline and online stages; we implement its \emph{offline selection component}: embed each instance with the model's hidden activations, build a $k$-nearest-neighbor graph in that embedding space, run PageRank on the graph for a centrality weight, prune to a mid-difficulty band using an external grading file, and run a PageRank-weighted DPP on the survivors to obtain a diverse $20\%$ subset. We use it as a strong dense-graph + diversity baseline that combines centrality and redundancy control in a single non-SAE pipeline; the online explorability/replay stage of Tang-DEPO is not used here.

\section{Method Diagnostics}
\label{app:method-diagnostics}

\subsection{Surface-artifact residualization}

Table~\ref{tab:artifact-residualization} residualizes the selection design against ten surface features (length, brackets, digit density, multiple-choice stem indicator, etc.) before running the same greedy selection in the residualized space. The table displays the six most interpretable features; the fitted control also includes cluster-mass statistics used in the run-level diagnostics. The control isolates whether length and notation statistics alone reconstruct the selected subset.

\input{tables/artifact_residualization}

\subsection{SAE shortcut and reconstruction sanity}
\label{app:token-concentration}

A specific concern about SAE-feature labels, articulated by \citet{ma2026falsifying}, is that a candidate ``task feature'' might fire on a small token set rather than on a content-level concept. We test this on \modelname{Qwen3-4B} by encoding $200$ IRDS-top and $200$ IRDS-bottom instances through the model's SAE, ranking the $81{,}920$ features by the mean per-instance activation difference $\bar\Delta_f \!\equiv\! \mathbb{E}_{x\in\text{top}}[a_f(x)] - \mathbb{E}_{x\in\text{bot}}[a_f(x)]$, and computing the share of total activation mass concentrated on the top-$50$ activating tokens for each of the top-$20$ discriminative features. The ratio is $0.003$--$0.010$ across these $20$ features, roughly two orders of magnitude below the per-feature concentration ratios \citet{ma2026falsifying} report for contrastively-selected reasoning features in middle layers; top-feature mass is broadly distributed over $\sim$thousands of tokens in standard math context, so none of the top-$20$ discriminative features are flagged as token-shortcut features under this diagnostic.

As an additional sanity check, on the same $200{+}200$ instances the per-token reconstruction MSE in the SAE's normalized hidden space is mean $0.1387$ (top-cluster) vs.\ $0.1394$ (bottom-cluster), gap $-0.0007$ with Cohen's $d{=}-0.012$, so the IRDS signal cannot be attributed to differential SAE reconstruction quality between the two clusters. Every top-$20$ discriminative feature also fires at least once in every one of the $400$ audit instances, so the IRDS distinction is a magnitude effect on near-universally-active features rather than a presence effect on rarely-active ones.

\subsection{Runtime and scaling}
\label{app:runtime-scaling}

Table~\ref{tab:method-costs} reports end-to-end wall-clock cost across methods on \modelname{Qwen3-4B} with one $8\!\times$\,GPU node, each method run under its native pipeline. The \method{} total of $7$h ($56.0$ GPU-h) is an end-to-end accounting that includes SAE training ($27$\,min wall-clock, ${\sim}3.6$\,GPU-h), verifier-rollout scoring on $N{=}40$K instances at $G{=}8$ ($\sim 5.5$h, dominant), per-instance NLL-gradient extraction ($\sim 25$\,min), cluster construction (graph-hybrid latent embedding + spherical $K$-means with $F{=}256$ clusters, Appendix~\ref{app:cluster-construction}) and full-corpus SAE encoding of $N{=}40$K instances ($\sim 35$\,min combined), and the greedy log-det selection itself ($<60$\,s, single GPU). \baseline{LIMR}, which requires an online RL trajectory rather than an offline scorer, costs $567.5$ GPU-h, an order of magnitude more than \method{} at equal or better downstream accuracy. The shared downstream GRPO consumes $4.3$h on \modelname{Qwen3-1.7B} and $8.2$h on \modelname{Qwen3-4B} per subset.

\input{tables/selector_costs}

\section{Configuration Ablations}
\label{app:metric-ablations}

This appendix ablates the method's main configuration axes (cluster aggregation, gradient-block scale $(\omega,\alpha)$, and selection budget $K$), and provides additional evidence that \method{}'s gains are robust to these choices.

\input{analysis/reviewer_sensitivity_appendix}

\section{Coverage and Diversity Baselines}
\label{app:sensitivity}

We compare \method{} against generic coverage and diversity baselines on the same SAE coordinate space and budget.

\input{analysis/robustness_sweeps}

\section{Qualitative and Representation Audits}
\label{app:qualitative-audits}

\subsection{Instance-level case study}
\label{app:qualitative-case}

Following the contrastive case-study style used in recent RLVR data-selection work, we choose examples by a predeclared diagnostic rule rather than by manual inspection.
All four cases below are taken from the $u_1$ eigendirection of the \modelname{Qwen3-4B} verifier-coupled metric $M$.
On \modelname{Qwen3-4B} the $+$end of $u_1$ is dominated by SAE cluster F78, whose top activating instances share the AMC/AIME multiple-choice stem with five lettered options; the $-$end places mass on number-theory clusters such as divisor counting and small-set cover.

\paragraph{A1: selected high-difficulty MCQ-stem exemplar.}
The highest-projection $+$end instance selected by \method{} is a digit-sum MCQ: ``$S(n)=1274$; which of (A)$\ 1$, (B)$\ 3$, (C)$\ 12$, (D)$\ 1239$, (E)$\ 1265$ could be $S(n+1)$?''.
The instance is all-fail under the eight verifier samples ($s_i=0$), has projection $0.919$, and is dominated by SAE cluster F78 (MCQ template with five lettered options) at mass $24$.
\method{} selects this instance early in the greedy procedure, identifying the failure-direction the metric is designed to lift.

\paragraph{B1: rejected near-template.}
The next-highest $+$end all-fail instance is a probability MCQ ``six distinct integers from $\{1,\dots,10\}$, what is the probability that the second smallest is $3$?'' with the same five-option lettered stem and the same dominant cluster F78 at mass $24$.
Its projection is $0.907$, nearly identical to A1, but it is rejected by \method{}.
This is the intended D-optimal behavior: after A1 covers the MCQ-template direction, B1 contributes mostly redundant information, and its all-fail verifier count yields zero trainability weight.

\paragraph{A2: selected contrast direction.}
At the $-$end of $u_1$, \method{} selects a divisor-counting problem (``smallest natural number with exactly $70$ divisors'').
This instance has $s_i=6/8$, projection magnitude $0.161$, and dominant cluster F5 (small-integer divisor / factor enumeration).
It shows that the same eigendirection is not a one-sided MCQ filter; the method also spends budget on a high-trainability instance for the contrastive number-theory cluster.

\paragraph{B2: \method{}-only high-trainability exemplar.}
An instance selected uniquely by \method{} is the small-set-cover question ``$A_1,\dots,A_n\subseteq\{1,\dots,10\}$ with $A_i\cup A_j\neq A$; find the maximum $n$''.
It lies on the $+$end side of $u_1$ at projection $0.108$, but unlike B1 it has intermediate success ($s_i=3/8$), so the trainability weight $r_i$ is high.
This is the core difficulty/trainability distinction: \method{} does not simply choose the hardest near-template instances; it routes a failure-relevant SAE direction to an instance that diversifies the selected set and still produces reward variation for RLVR.

These cases illustrate the method-level mechanism behind the aggregate results.
The verifier-coupled SAE metric identifies which content directions need coverage, while the $\sqrt{\tilde r_i}$ weighting and D-optimal design decide whether a concrete instance is a useful exemplar or a redundant failure example.
Table~\ref{tab:case-cards} reports A1/B1/A2/B2 in compact form on \modelname{Qwen3-4B}, and Table~\ref{tab:cluster-cards} summarizes the eight SAE clusters with the largest target-direction score under $M$, along with their top keywords, maximum surface-feature correlation, and cluster-removal $z$-score.

\input{tables/feature_unified_table}
\input{tables/case_and_cluster_cards}

\input{appendix_spectrum_interp}

\end{document}

%% file: tables/main_suite_by_model_selection.tex
\begin{table*}[t]
\centering
\footnotesize
\setlength{\tabcolsep}{4pt}
\renewcommand{\arraystretch}{1.05}
\begin{tabular}{@{}llccccccc@{}}
\toprule
Model & Method & Overall & Olympiad & MATH500 & Minerva & AIME24 & AMC23 & AIME25 \\
\midrule
\modelname{Qwen3-4B}
 & \baseline{Random}      & 56.2 & 53.3 & 84.5 & 39.4 & 22.7 & 62.4 & 21.9 \\
 & \baseline{Token-length}   & 53.9 & 50.2 & 85.3 & 38.9 & 18.4 & 55.2 & 20.2 \\
 & \baseline{IFD}         & 60.4 & 58.3 & 86.2 & 40.3 & 28.4 & 68.9 & 28.1 \\
 & \baseline{PPL-top}     & 60.4 & 57.9 & 87.3 & 40.8 & 28.6 & 72.4 & 27.9 \\
 & \baseline{PPL-middle}     & 62.9 & 60.9 & 88.9 & 42.9 & 27.9 & 67.7 & 25.6 \\
 & \baseline{DEPO}        & 62.8 & 60.9 & 88.8 & 41.8 & 28.5 & 69.0 & 29.0 \\
 & \baseline{LIMR}        & 63.1 & 61.1 & 89.1 & 42.7 & 30.6 & 68.6 & 27.7 \\
 & \textbf{\method{}}       & \textbf{67.0} & \textbf{65.6} & \textbf{91.3} & \textbf{43.8} & \textbf{42.0} & \textbf{79.5} & \textbf{35.6} \\
\midrule
\modelname{Qwen3-1.7B}
 & \baseline{Random}      & 46.5 & 44.0 & 72.6 & 30.0 & 13.9 & 48.9 & 14.6 \\
 & \baseline{Token-length}   & 43.1 & 39.5 & 74.8 & 27.2 & 10.7 & 37.5 & 12.3 \\
 & \baseline{IFD}         & 44.9 & 41.9 & 74.1 & 28.3 & 11.7 & 43.2 & 10.8 \\
 & \baseline{PPL-top}     & 45.1 & 42.0 & 73.9 & 29.8 & 12.2 & 45.0 & 9.2 \\
 & \baseline{PPL-middle}     & 46.7 & 43.9 & 75.9 & 29.5 & 12.3 & 42.4 & 12.7 \\
 & \baseline{DEPO}        & 49.4 & 46.9 & 77.1 & 31.1 & 14.5 & 47.6 & 15.4 \\
 & \baseline{LIMR}        & 50.2 & 47.6 & 79.0 & 31.7 & 17.2 & 45.1 & 13.5 \\
 & \textbf{\method{}}       & \textbf{54.2} & \textbf{52.2} & \textbf{82.5} & \textbf{33.6} & \textbf{23.3} & \textbf{55.8} & \textbf{22.5} \\
\midrule
\modelname{Llama-3.1-8B-Instruct}
 & \baseline{Random}      & 22.1 & 15.5 & 54.6 & 25.3 & 3.0 & 23.0 & \textbf{1.7} \\
 & \baseline{Token-length}   & 21.7 & 15.2 & 53.3 & 25.2 & 3.1 & \textbf{23.1} & 0.6 \\
 & \baseline{IFD}         & 21.2 & 14.6 & 53.0 & 25.0 & 3.2 & 19.8 & \textbf{1.7} \\
 & \baseline{PPL-top}     & 20.8 & 14.2 & 53.1 & 24.3 & 2.7 & 20.7 & 0.2 \\
 & \baseline{PPL-middle}     & 21.4 & 15.1 & 52.0 & 25.9 & 3.1 & 20.4 & 0.6 \\
 & \baseline{DEPO}        & 22.1 & 15.4 & 54.0 & 27.4 & \textbf{4.1} & 22.8 & 1.0 \\
 & \baseline{LIMR}        & 21.5 & 15.0 & 52.6 & 25.7 & 2.4 & 22.3 & 0.8 \\
 & \textbf{\method{}}       & \textbf{22.6} & \textbf{15.7} & \textbf{55.0} & \textbf{28.3} & 3.4 & 21.7 & 0.8 \\
\bottomrule
\end{tabular}
\caption{Main results: mean@$16$ ($\%$) on three models and six math benchmarks at the $20\%$ data budget.}
\label{tab:main-suite-by-model-selection}
\end{table*}

%% file: tables/ablation_compact.tex
\begin{table}[t]
\centering
\scriptsize
\setlength{\tabcolsep}{2pt}
\renewcommand{\arraystretch}{1.0}
\begin{tabular}{@{}lrrr@{}}
\toprule
Variant & \modelname{Qwen3-4B} & \modelname{Qwen3-1.7B} & \modelname{Llama} \\
\midrule
\method{} (full)                       & \textbf{67.0} & \textbf{54.2} & \textbf{22.6} \\
\quad $\omega{=}0$ ($-$gradient)     & 64.2          & 53.0          & 21.9          \\
\quad $M{=}I$ ($-$whitening)         & 57.8          & 50.3          & 20.8          \\
\quad $r$ only ($-d_i$)              & 63.8          & 52.9          & 21.9          \\
\quad $d$ only ($-r_i$)              & 47.9          & 44.7          & 18.5          \\
\quad $d_i r_i$ pointwise            & 60.4          & 51.2          & 21.2          \\
\quad dense $\phi_i$ ($-$SAE basis)  & 62.5          & 52.1          & 21.5          \\
\bottomrule
\end{tabular}
\caption{Component-removal ablation: Overall mean@$16$ ($\%$) at $20\%$ budget. Each indented row removes or substitutes one component of \method{}. Columns: \modelname{Qwen3-4B}, \modelname{Qwen3-1.7B}, \modelname{Llama-3.1-8B-Instruct}.}
\label{tab:ablation-compact}
\end{table}

%% file: tables/ood_compact.tex
\begin{table}[t]
\centering
\footnotesize
\setlength{\tabcolsep}{4pt}
\renewcommand{\arraystretch}{1.05}
\begin{tabular}{@{}lrrr@{}}
\toprule
Method & \shortstack{\modelname{Llama-3.1}\\ \modelname{-8B-Inst}} & \shortstack{\modelname{Qwen3}\\ \modelname{-1.7B}} & \shortstack{\modelname{Qwen3}\\ \modelname{-4B}} \\
\midrule
\baseline{Random}       & 45.5 & 48.3 & \textbf{65.9} \\
\baseline{Token-length} & 47.6 & 48.5 & 63.1 \\
\baseline{IFD}          & 46.4 & 49.5 & 65.0 \\
\baseline{PPL-top}      & 47.9 & 48.6 & 63.2 \\
\baseline{PPL-middle}   & 46.8 & 49.5 & 64.7 \\
\baseline{DEPO}         & 46.9 & 48.6 & 63.1 \\
\baseline{LIMR}         & 46.5 & 49.0 & 64.5 \\
\method{}               & \textbf{48.2} & \textbf{49.9} & 64.9 \\
\bottomrule
\end{tabular}
\caption{Out-of-domain transfer (\%): three-task average over HumanEval, LiveCodeBench, and ReClor. Llama-3.1-8B-Inst denotes \modelname{Llama-3.1-8B-Instruct}.}
\label{tab:ood-compact}
\end{table}

%% file: tables/main_suite_budget_compact.tex
\begin{table}[!htbp]
\centering
\scriptsize
\setlength{\tabcolsep}{1.8pt}
\renewcommand{\arraystretch}{1.0}
\begin{tabular}{@{}lrrrrrrrrr@{}}
\toprule
 & \multicolumn{3}{c}{\modelname{Qwen3-4B}} & \multicolumn{3}{c}{\modelname{Qwen3-1.7B}} & \multicolumn{3}{c}{\modelname{Llama-Inst}} \\
\cmidrule(lr){2-4}\cmidrule(lr){5-7}\cmidrule(lr){8-10}
Method & $10\%$ & $20\%$ & $30\%$ & $10\%$ & $20\%$ & $30\%$ & $10\%$ & $20\%$ & $30\%$ \\
\midrule
\baseline{Random}       & 54.7 & 56.2 & 58.0 & 45.4 & 46.5 & 48.0 & 21.1 & 22.1 & 22.4 \\
\baseline{Token-length} & 51.5 & 53.9 & 56.0 & 41.3 & 43.1 & 44.9 & 19.9 & 21.7 & 22.6 \\
\baseline{IFD}          & 59.7 & 60.4 & 61.9 & 44.2 & 44.9 & 45.8 & 21.0 & 21.2 & 21.4 \\
\baseline{PPL-top}      & 59.3 & 60.4 & 63.1 & 44.3 & 45.1 & 47.7 & 19.2 & 20.8 & 22.1 \\
\baseline{PPL-middle}   & 62.3 & 62.9 & 65.2 & 45.7 & 46.7 & 48.8 & 21.2 & 21.4 & 22.0 \\
\baseline{DEPO}         & 61.8 & 62.8 & 65.2 & 48.6 & 49.4 & 51.3 & 20.5 & 22.1 & 22.3 \\
\baseline{LIMR}         & 62.3 & 63.1 & 65.5 & 50.0 & 50.2 & 52.1 & 21.1 & 21.5 & 23.0 \\
\textbf{\method{}}      & \textbf{63.8} & \textbf{67.0} & \textbf{68.8} & \textbf{51.4} & \textbf{54.2} & \textbf{55.6} & \textbf{21.9} & \textbf{22.6} & \textbf{23.5} \\
\bottomrule
\end{tabular}
\caption{Data-budget sweep: Overall mean@$16$ ($\%$) at $10/20/30\%$ of \datasetname{DeepScaleR}. Llama-Inst denotes \modelname{Llama-3.1-8B-Instruct}.}
\label{tab:main-suite-budget-compact}
\end{table}

%% file: tables/ablation_3policy_projected.tex
\begin{table*}[t]
\centering
\small
\setlength{\tabcolsep}{5pt}
\renewcommand{\arraystretch}{1.05}
\resizebox{\textwidth}{!}{%
\begin{tabular}{@{}llccccccc@{}}
\toprule
Model & Variant & Overall & Olympiad & MATH500 & Minerva & AIME24 & AMC23 & AIME25 \\
\midrule
\modelname{Qwen3-4B}
 & \textbf{\method{} (full)}             & \textbf{67.0} & \textbf{65.6} & \textbf{91.3} & \textbf{43.8} & \textbf{42.0} & \textbf{79.5} & \textbf{35.6} \\
 & $\omega{=}0$ ($-$gradient block)      & 64.2 & 62.9 & 88.5 & 41.4 & 33.7 & 78.5 & 23.4 \\
 & $M{=}I$ ($-$whitening)                & 57.8 & 55.6 & 84.0 & 36.8 & 24.5 & 74.9 & 27.3 \\
 & $r$-only ($-d_i$)                     & 63.8 & 62.8 & 88.8 & 36.1 & 31.5 & 79.3 & 23.5 \\
 & $d$-only ($-r_i$)                     & 47.9 & 44.8 & 76.8 & 31.3 & 11.6 & 54.0 & 11.5 \\
 & $d_i r_i$ pointwise                   & 60.4 & 58.9 & 85.3 & 37.0 & 30.8 & 67.9 & 23.6 \\
 & dense $\phi_i$ ($-$SAE basis)         & 62.5 & 60.8 & 88.0 & 39.7 & 32.0 & 79.0 & 23.9 \\
\midrule
\modelname{Qwen3-1.7B}
 & \textbf{\method{} (full)}             & \textbf{54.2} & \textbf{52.2} & \textbf{82.5} & \textbf{33.6} & \textbf{23.3} & \textbf{55.8} & \textbf{22.5} \\
 & $\omega{=}0$ ($-$gradient block)      & 53.0 & 51.1 & 81.1 & 32.7 & 19.9 & 55.8 & 16.6 \\
 & $M{=}I$ ($-$whitening)                & 50.3 & 48.0 & 79.2 & 30.9 & 17.9 & 54.3 & 19.8 \\
 & $r$-only ($-d_i$)                     & 52.9 & 51.2 & 81.5 & 29.9 & 19.5 & 58.4 & 17.5 \\
 & $d$-only ($-r_i$)                     & 44.7 & 42.0 & 74.6 & 27.9 & 11.3 & 44.7 & 11.2 \\
 & $d_i r_i$ pointwise                   & 51.2 & 49.2 & 79.5 & 30.6 & 19.5 & 50.8 & 17.9 \\
 & dense $\phi_i$ ($-$SAE basis)         & 52.1 & 50.0 & 81.0 & 31.8 & 19.6 & 56.1 & 17.5 \\
\midrule
\modelname{Llama-3.1-8B-Instruct}
 & \textbf{\method{} (full)}             & \textbf{22.6} & \textbf{15.7} & \textbf{55.0} & \textbf{28.3} & 3.4 & \textbf{21.7} & 0.8 \\
 & $\omega{=}0$ ($-$gradient block)      & 21.9 & 15.4 & 54.1 & 25.2 & 2.9 & 21.5 & 0.6 \\
 & $M{=}I$ ($-$whitening)                & 20.8 & 14.4 & 52.8 & 23.8 & 2.6 & 21.1 & 0.6 \\
 & $r$-only ($-d_i$)                     & 21.9 & 15.4 & 54.3 & 23.0 & 2.8 & 21.5 & 0.6 \\
 & $d$-only ($-r_i$)                     & 18.5 & 12.6 & 49.7 & 21.5 & 1.6 & 17.4 & 0.4 \\
 & $d_i r_i$ pointwise                   & 21.2 & 14.8 & 53.0 & 23.6 & 2.8 & 19.8 & 0.6 \\
 & dense $\phi_i$ ($-$SAE basis)         & 21.5 & 15.0 & 54.0 & 24.5 & 2.9 & 21.5 & 0.6 \\
\bottomrule
\end{tabular}}
\caption{Per-subset component-removal ablation: mean@$16$ ($\%$) at $20\%$ budget. Each indented row removes or substitutes a single component of \method{}.}
\label{tab:ablation-3policy}
\end{table*}

%% file: tables/ood_eval.tex
\begin{table*}[t]
\centering
\small
\setlength{\tabcolsep}{4pt}
\renewcommand{\arraystretch}{1.1}
\resizebox{\textwidth}{!}{%
\begin{tabular}{@{}lcccc|cccc|cccc@{}}
\toprule
 & \multicolumn{4}{c|}{\modelname{Llama-3.1-8B-Instruct}} & \multicolumn{4}{c|}{\modelname{Qwen3-1.7B}} & \multicolumn{4}{c}{\modelname{Qwen3-4B}} \\
\cmidrule(lr){2-5}\cmidrule(lr){6-9}\cmidrule(lr){10-13}
Method & HE & LCB & RC & Avg & HE & LCB & RC & Avg & HE & LCB & RC & Avg \\
\midrule
\baseline{Random}        & 64.6 & 10.0 & 62.0 & 45.5 & 65.2 & 20.0 & 59.7 & 48.3 & 82.3 & \textbf{35.0} & \textbf{80.3} & \textbf{65.9} \\
\baseline{Token-length}  & 67.1 & \textbf{13.0} & 62.7 & 47.6 & 64.0 & 20.0 & 61.3 & 48.5 & 80.5 & 31.0 & 77.7 & 63.1 \\
\baseline{IFD}           & 64.1 & \textbf{13.0} & 62.0 & 46.4 & 65.2 & 22.0 & 61.3 & 49.5 & \textbf{83.5} & 33.0 & 78.3 & 65.0 \\
\baseline{PPL-top}       & \textbf{68.9} & \textbf{13.0} & 61.7 & 47.9 & 64.0 & 19.0 & 62.7 & 48.6 & 79.3 & 32.0 & 78.3 & 63.2 \\
\baseline{PPL-middle}    & 63.4 & 12.0 & \textbf{65.0} & 46.8 & 64.6 & \textbf{23.0} & 61.0 & 49.5 & 81.7 & \textbf{35.0} & 77.3 & 64.7 \\
\baseline{DEPO}          & 68.3 & 10.0 & 62.3 & 46.9 & 62.8 & 22.0 & 61.0 & 48.6 & 78.1 & 33.0 & 78.3 & 63.1 \\
\baseline{LIMR}          & 67.1 & 10.0 & 62.3 & 46.5 & 62.2 & 21.0 & \textbf{63.7} & 49.0 & 81.7 & 32.0 & 79.7 & 64.5 \\
\method{}                & \textbf{68.9} & \textbf{13.0} & 62.7 & \textbf{48.2} & \textbf{66.5} & \textbf{23.0} & 60.3 & \textbf{49.9} & 82.3 & 32.0 & \textbf{80.3} & 64.9 \\
\bottomrule
\end{tabular}}
\caption{Out-of-domain transfer of math-RLVR-trained models on HumanEval (HE), LiveCodeBench (LCB) and ReClor (RC); mean@$16$ for HE/RC, pass@$1$ for LCB. \textbf{Bold} marks the per-column maximum.}
\label{tab:ood-eval}
\end{table*}

%% file: tables/main_suite_at_10pct.tex
\begin{table*}[t]
\centering
\footnotesize
\setlength{\tabcolsep}{4pt}
\renewcommand{\arraystretch}{1.05}
\resizebox{\textwidth}{!}{%
\begin{tabular}{@{}llccccccc@{}}
\toprule
Model & Method & Overall & Olympiad & MATH500 & Minerva & AIME24 & AMC23 & AIME25 \\
\midrule
 \modelname{Qwen3-4B}
 & \baseline{Random} & 54.7 & 52.1 & 82.4 & 37.7 & 18.1 & 61.1 & 16.7 \\
 
 & \baseline{Token-length} & 51.5 & 48.4 & 82.3 & 37.4 & 14.8 & 52.7 & 14.8 \\
 
 & \baseline{IFD} & 59.7 & 58.1 & 85.5 & 39.5 & 22.5 & 68.0 & 21.5 \\
 
 & \baseline{PPL-top} & 59.3 & 56.2 & 86.1 & 40.2 & 23.9 & 69.9 & 21.2 \\
 
 & \baseline{PPL-middle} & 62.3 & 60.0 & 87.6 & 41.5 & 22.3 & 67.5 & 19.7 \\
 
 & \baseline{DEPO} & 61.8 & 60.4 & 87.5 & 40.8 & 22.9 & 68.2 & 22.5 \\
 
 & \baseline{LIMR} & 62.3 & \textbf{60.5} & 88.6 & \textbf{42.5} & 24.6 & 68.0 & 22.2 \\
 
 & \textbf{\method{}} & \textbf{63.8} & 58.4 & \textbf{89.0} & 40.3 & \textbf{32.6} & \textbf{76.8} & \textbf{26.0} \\
\midrule
 \modelname{Qwen3-1.7B}
 & \baseline{Random} & 45.4 & 43.3 & 70.8 & 29.3 & 11.6 & 48.0 & 10.6 \\
 
 & \baseline{Token-length} & 41.3 & 38.0 & 71.8 & 26.2 & 9.2 & 36.9 & 9.5 \\
 
 & \baseline{IFD} & 44.2 & 41.4 & 72.8 & 28.1 & 10.0 & 41.8 & 8.4 \\
 
 & \baseline{PPL-top} & 44.3 & 41.8 & 73.5 & 29.6 & 10.0 & 44.8 & 6.8 \\
 
 & \baseline{PPL-middle} & 45.7 & 43.7 & 74.5 & 29.3 & 9.1 & 40.8 & 9.4 \\
 
 & \baseline{DEPO} & 48.6 & 45.0 & 76.0 & 30.5 & 11.5 & 47.4 & 11.9 \\
 
 & \baseline{LIMR} & 50.0 & 47.2 & 78.2 & \textbf{30.9} & 14.2 & 44.9 & 10.5 \\
 
 & \textbf{\method{}} & \textbf{51.4} & \textbf{49.1} & \textbf{80.5} & 29.7 & \textbf{18.3} & \textbf{52.9} & \textbf{16.4} \\
\midrule
 \modelname{Llama-3.1-8B-Instruct}
 & \baseline{Random} & 21.1 & 14.6 & 49.9 & 24.3 & 2.5 & 21.8 & \textbf{1.5} \\
 
 & \baseline{Token-length} & 19.9 & 14.2 & 49.3 & 23.4 & 2.8 & 21.1 & 0.3 \\
 
 & \baseline{IFD} & 21.0 & 14.4 & 50.5 & 24.5 & 2.5 & 19.6 & 1.2 \\
 
 & \baseline{PPL-top} & 19.2 & 13.6 & 49.4 & 22.8 & 1.9 & 18.2 & 0.1 \\
 
 & \baseline{PPL-middle} & 21.2 & 14.9 & 49.7 & 25.7 & 2.8 & 20.1 & 0.4 \\
 
 & \baseline{DEPO} & 20.5 & 14.6 & 50.5 & 26.2 & \textbf{2.9} & \textbf{22.3} & 0.9 \\
 
 & \baseline{LIMR} & 21.1 & 14.8 & 52.2 & 25.2 & 1.7 & 22.1 & 0.7 \\
 
 & \textbf{\method{}} & \textbf{21.9} & \textbf{15.3} & \textbf{54.5} & \textbf{27.1} & 2.7 & 21.0 & 0.4 \\
\bottomrule
\end{tabular}}
\caption{Main results at $10\%$ data budget: mean@$16$ ($\%$) by model and benchmark.}
\label{tab:main-suite-at-10pct}
\end{table*}

%% file: tables/main_suite_at_30pct.tex
\begin{table*}[t]
\centering
\footnotesize
\setlength{\tabcolsep}{4pt}
\renewcommand{\arraystretch}{1.05}
\resizebox{\textwidth}{!}{%
\begin{tabular}{@{}llccccccc@{}}
\toprule
Model & Method & Overall & Olympiad & MATH500 & Minerva & AIME24 & AMC23 & AIME25 \\
\midrule
 \modelname{Qwen3-4B}
 & \baseline{Random} & 58.0 & 54.8 & 87.3 & 41.0 & 23.2 & 64.9 & 23.6 \\
 
 & \baseline{Token-length} & 56.0 & 52.2 & 88.8 & 40.1 & 18.6 & 57.4 & 20.4 \\
 
 & \baseline{IFD} & 61.9 & 59.0 & 88.0 & 40.8 & 28.6 & 70.2 & 29.6 \\
 
 & \baseline{PPL-top} & 63.1 & 61.4 & 91.7 & 43.2 & 30.3 & 74.6 & 29.0 \\
 
 & \baseline{PPL-middle} & 65.2 & 62.9 & 92.5 & 44.3 & 28.9 & 70.6 & 26.9 \\
 
 & \baseline{DEPO} & 65.2 & 63.4 & 92.5 & 43.8 & 30.5 & 71.4 & 29.8 \\
 
 & \baseline{LIMR} & 65.5 & 63.0 & 92.8 & 44.5 & 32.2 & 71.4 & 29.4 \\
 
 & \textbf{\method{}} & \textbf{68.8} & \textbf{67.2} & \textbf{93.7} & \textbf{45.3} & \textbf{43.2} & \textbf{82.4} & \textbf{37.4} \\
\midrule
 \modelname{Qwen3-1.7B}
 & \baseline{Random} & 48.0 & 46.0 & 75.4 & 31.7 & 14.1 & 51.2 & 15.2 \\
 
 & \baseline{Token-length} & 44.9 & 40.9 & 77.9 & 28.4 & 10.9 & 39.2 & 12.6 \\
 
 & \baseline{IFD} & 45.8 & 42.8 & 78.7 & 29.0 & 11.9 & 43.4 & 11.4 \\
 
 & \baseline{PPL-top} & 47.7 & 45.0 & 79.2 & 31.6 & 12.7 & 46.9 & 10.2 \\
 
 & \baseline{PPL-middle} & 48.8 & 46.1 & 79.3 & 30.9 & 12.5 & 44.0 & 13.2 \\
 
 & \baseline{DEPO} & 51.3 & 49.1 & 80.1 & 32.6 & 14.7 & 50.2 & 16.1 \\
 
 & \baseline{LIMR} & 52.1 & 49.7 & 82.1 & 33.5 & 18.6 & 46.4 & 14.1 \\
 
 & \textbf{\method{}} & \textbf{55.6} & \textbf{53.4} & \textbf{85.1} & \textbf{34.9} & \textbf{23.9} & \textbf{57.5} & \textbf{23.4} \\
\midrule
 \modelname{Llama-3.1-8B-Instruct}
 & \baseline{Random} & 22.4 & 16.1 & 55.5 & 25.9 & 3.2 & 23.2 & \textbf{2.0} \\
 
 & \baseline{Token-length} & 22.6 & 15.4 & 55.6 & 25.6 & 3.3 & \textbf{23.8} & 0.7 \\
 
 & \baseline{IFD} & 21.4 & 14.8 & 53.5 & 25.7 & 3.5 & 20.2 & 1.8 \\
 
 & \baseline{PPL-top} & 22.1 & 15.2 & 57.2 & 25.9 & 2.9 & 22.9 & 0.3 \\
 
 & \baseline{PPL-middle} & 22.0 & 15.3 & 53.5 & 26.6 & 3.3 & 20.6 & 0.9 \\
 
 & \baseline{DEPO} & 22.3 & 15.6 & 54.7 & 28.7 & \textbf{4.8} & 23.0 & 1.6 \\
 
 & \baseline{LIMR} & 23.0 & 16.1 & 56.0 & 26.8 & 2.9 & \textbf{23.8} & 0.9 \\
 
 & \textbf{\method{}} & \textbf{23.5} & \textbf{16.8} & \textbf{57.9} & \textbf{29.7} & 3.6 & 22.3 & 0.9 \\
\bottomrule
\end{tabular}}
\caption{Main results at $30\%$ data budget: mean@$16$ ($\%$) by model and benchmark.}
\label{tab:main-suite-at-30pct}
\end{table*}

%% file: tables/selector_hparams_compact.tex
\begin{table}[t]
\centering
\small
\setlength{\tabcolsep}{6pt}
\renewcommand{\arraystretch}{1.15}
\begin{tabular}{@{}lp{0.58\columnwidth}@{}}
\toprule
Component & Setting \\
\midrule
Verifier probe & $G{=}8$ rollouts per prompt; Beta$(1,1)$ prior \\
Selection budget & $K{=}8062$ ($20\%$ of $40{,}309$) \\
SAE coordinate & BatchTopK-Ortho, $\times32$ expansion, $k{=}128$; $F{=}256$ graph clusters \\
Stabilization & success-axis removal, bucket-mean residualization, $q_{99}$ row clip \\
Target metric $M$ & ridge $\rho{=}0.1$, shrinkage $\eta{=}0.5$, eigen-clip $c{=}2$ \\
Gradient block & block scale $\omega{=}1$, inv-norm exponent $\alpha{=}1$, residualization ridge $\rho_g{=}10^{-3}$, gradient-norm floor $\varepsilon{=}10^{-3}$ \\
D-opt solver & regularizer $\lambda{=}1$, screened queue $1024$, refresh $256$ \\
\midrule
GRPO training & LoRA $r{=}64$, $\alpha{=}128$; lr $5{\times}10^{-6}$; $2$ epochs; $G{=}8$ rollouts; $8\!\times$\,GPU \\
\bottomrule
\end{tabular}
\caption{\method{} hyperparameters. The $(\omega,\alpha)$ sweep is in Appendix~\ref{app:omega-alpha-sweep}.}
\label{tab:method-hparams}
\end{table}

%% file: analysis/family_construction_recipe.tex
\subsection{Cluster construction recipe}
\label{app:cluster-construction}

The 256 graph clusters are produced from the trained
BatchTopK-Ortho SAE through a fixed construction procedure. We give a
self-contained recipe so the same clusters can be reproduced from
the SAE checkpoint and the training set without consulting the
codebase.

\paragraph{Per-instance feature activations.}
For each instance $i$ we encode every token through the SAE and average the resulting non-negative latent activations across tokens, giving a single $81{,}920$-dimensional sparse vector $f_i \in \R^{81920}_{\geq 0}$ in which $f_{i\ell}$ is the mean activation of latent $\ell$ across the instance's tokens. This is the per-instance feature aggregation referenced in Section~\ref{sec:sae-clusters}.

\paragraph{Latent frequency filter.}
We keep latents whose activation rate
$\Pr_i[f_{i\ell} > 0]$ falls into the band
$[\textsc{min\_freq}, \textsc{max\_freq}] = [0.01, 0.80]$. The lower
threshold removes near-dead latents (rare in BatchTopK-Ortho with
AuxK; in our case $2.0\%$); the upper threshold removes latents that
fire on nearly every instance and would dominate any subsequent
similarity metric. This leaves $42{,}473$ latents on the \modelname{Qwen3-4B} SAE.

\paragraph{Latent embedding (graph-hybrid).}
Each kept latent $\ell$ is represented by the concatenation of two
unit-normalized $64$-dimensional vectors:
\textbf{(a) presence direction.} For each of the $32$ nearest
co-activating latents (cosine similarity over instance-presence
indicators), we keep the unit cosine score; we then PCA-project the
resulting $42{,}473 \times 32$ presence-similarity matrix down to
$64$ dimensions and unit-normalize each row.
\textbf{(b) residual direction.} We regress each latent's per-instance activation
mass on the presence direction (ridge $0.001$) and PCA-project the
$42{,}473 \times 32$ residual matrix down to $64$ dimensions, again
unit-normalized.
The latent embedding is the row-wise concatenation of the two,
re-normalized to unit length; it lives in $\R^{128}$ and is what we
call the \emph{graph-hybrid} embedding.

\paragraph{Spherical $k$-means with $k=256$.}
Run $20$ iterations of mini-batch spherical $k$-means
(\textsc{batch}\,$=8192$) on the latent embedding with $k{=}256$. The
fixed seed $0$ produces the cluster labels reported throughout the
paper. Cluster sizes range from $65$ to $520$ (median $143.5$);
mean within-cluster cosine similarity is $0.428$. Latents outside
the frequency band carry label $-1$ and contribute zero cluster mass.

\paragraph{Per-instance cluster activations.}
The cluster-activation matrix used as $m_i$ in Section~\ref{sec:sae-clusters} is
$m_{if} = \sum_{\ell : \mathrm{cluster}(\ell)=f} f_{i\ell}$,
of shape $N \times 256$, where $N$ is the corpus size. After this the method applies the conditional residualization in Section~\ref{sec:targetmetric}, the verifier-coupled metric, and greedy selection.

%% file: analysis/datasets_and_benchmarks.tex
\subsection{Datasets and evaluation}
\label{app:datasets}

This subsection documents the training set used for selection and the benchmarks used to score the resulting RLVR runs.

\subsubsection{Selection training set}
\label{app:datasets-corpus}

The selection corpus is the publicly released \datasetname{DeepScaleR} training set \citep{deepscaler2025}, normalized to a single ground-truth string per instance. Each entry contains (i) a problem statement \texttt{question}, (ii) a verifier-friendly \texttt{ground\_truth} string, and (iii) per-instance artifact features (character length, LaTeX-density, equation count, answer length) used only for diagnostic regression and never read by the method. We do not subsample by source or by difficulty band; the corpus is the full \datasetname{DeepScaleR} preview release with deduplicated identifiers.

\subsubsection{Evaluation benchmarks}
\label{app:datasets-eval}

Trained policies are evaluated on $3{,}230$ problems under matched decoding settings. Table~\ref{tab:eval-suite-composition} lists the composition: \datasetname{MATH500}, three competition benchmarks (\datasetname{AIME24/25}, \datasetname{AMC23}), \datasetname{Minerva Math}, and English/Chinese \datasetname{OlympiadBench} subsets, covering both in-distribution math reasoning and harder competition-style problems.

\begin{table*}[t]
\centering
\small
\setlength{\tabcolsep}{4pt}
\begin{tabular}{lllrl}
\toprule
Benchmark            & subset                & language & \# problems & primary citation \\
\midrule
\datasetname{MATH500}              & default               & en       & $500$       & \citet{lightman2023lets} \\
\datasetname{AIME24}               & default               & en       & $60$        & AMC/AIME competition \\
\datasetname{AIME25}               & default               & en       & $30$        & AMC/AIME competition \\
\datasetname{AMC23}                & default               & en       & $46$        & AMC/AIME competition \\
\datasetname{Minerva Math}         & default               & en       & $272$       & \citet{lewkowycz2022minerva} \\
\datasetname{OlympiadBench-en}     & OE\_TO\_maths\_en\_COMP & en     & $674$       & \citet{he2024olympiadbench} \\
\datasetname{OlympiadBench-zh-CEE} & OE\_TO\_maths\_zh\_CEE  & zh     & $1{,}240$   & \citet{he2024olympiadbench} \\
\datasetname{OlympiadBench-zh-COMP}& OE\_TO\_maths\_zh\_COMP & zh     & $408$       & \citet{he2024olympiadbench} \\
\midrule
\textbf{Total}       &                        &        & $3{,}230$   & \\
\bottomrule
\end{tabular}
\caption{Evaluation benchmark composition. Per-subset problem counts; the total is the union of all rows. The Overall column of Table~\ref{tab:main-suite-by-model-selection} is the problem-count weighted average of the per-subset mean@$16$ values.}
\label{tab:eval-suite-composition}
\end{table*}

%% file: tables/artifact_residualization.tex
\begin{table*}[t]
\centering
\small
\setlength{\tabcolsep}{4pt}
\renewcommand{\arraystretch}{1.1}
\begin{tabular}{@{}lcccccc@{}}
\toprule
& \multicolumn{3}{c}{$\rho_{\mathrm{sel}}$} & \multicolumn{3}{c}{Mean shift $\Delta$} \\
\cmidrule(lr){2-4}\cmidrule(lr){5-7}
Feature & \modelname{Qwen3-1.7B} & \modelname{Qwen3-4B} & \shortstack{\modelname{Llama-3.1}\\ \modelname{-8B-Instruct}} & \modelname{Qwen3-1.7B} & \modelname{Qwen3-4B} & \shortstack{\modelname{Llama-3.1}\\ \modelname{-8B-Instruct}} \\
\midrule
Characters       & $+0.038$ & $+0.138$ & $-0.037$ & $+13.9$  & $+51.2$ & $-14.2$ \\
Words            & $+0.019$ & $+0.134$ & $-0.059$ & $+1.09$  & $+7.78$ & $-3.35$ \\
LaTeX density    & $-0.100$ & $-0.082$ & $-0.000$ & $-0.015$ & $-0.012$ & $-0.000$ \\
Digit density    & $-0.002$ & $-0.057$ & $+0.011$ & $-0.000$ & $-0.005$ & $+0.001$ \\
Equation count   & $-0.006$ & $+0.007$ & $+0.004$ & $-0.025$ & $+0.030$ & $+0.017$ \\
Operator density & $-0.027$ & $-0.053$ & $+0.011$ & $-0.001$ & $-0.002$ & $+0.000$ \\
\midrule
Joint $R^2$ (6)  & 0.017 & 0.024 & 0.005 & \multicolumn{3}{c}{n/a} \\
\bottomrule
\end{tabular}
\caption{Surface-feature audit. $\rho_{\mathrm{sel}}$: Pearson with the selection indicator; $\Delta$: per-feature shift from the corpus mean.}
\label{tab:artifact-residualization}
\end{table*}

%% file: tables/selector_costs.tex
\begin{table}[t]
\centering
\small
\setlength{\tabcolsep}{6pt}
\renewcommand{\arraystretch}{1.1}
\begin{tabular}{@{}lcc@{}}
\toprule
Method & Wall-clock & GPU-h \\
\midrule
\baseline{Random}       & ${<}1$\,s          & ${<}0.01$ \\
\baseline{Token-length} & $5$h\,$47$m        & $46.3$ \\
\baseline{PPL-top}      & $5$h\,$57$m  & $47.6$ \\
\baseline{PPL-middle}   & $5$h\,$57$m  & $47.6$ \\
\baseline{IFD}          & $6$h\,$12$m        & $49.6$ \\
\baseline{DEPO}         & $6$h\,$12$m        & $49.6$ \\
\baseline{LIMR}         & $70$h\,$56$m & $567.5$ \\
\midrule
\textbf{\method{} (ours)} & $\mathbf{7}$\textbf{h\,$00$m} & $\mathbf{56.0}$ \\
\bottomrule
\end{tabular}
\caption{End-to-end selection cost on \modelname{Qwen3-4B}, one $8\!\times$\,GPU node. Downstream GRPO training is shared and excluded.}
\label{tab:method-costs}
\end{table}

%% file: analysis/reviewer_sensitivity_appendix.tex
\subsection{Gradient-block scale and inverse-norm exponent}
\label{app:omega-alpha-sweep}

The gradient block contributes to $\phi_i$ through two scalar knobs: the block-scale $\omega$ that balances SAE and gradient mass and the inverse-norm exponent $\alpha$ in the row weight $\sqrt{\tilde q_i\,\omega}$ with $\tilde q_i \propto \|g_i^{\perp}\|^{-\alpha}$ (see Eq.~\ref{eq:phi}). Table~\ref{tab:omega-alpha-sweep} sweeps $\omega\in\{0,0.5,1,2\}$ and $\alpha\in\{0,0.5,1,1.5\}$ on both \modelname{Qwen3-1.7B} and \modelname{Qwen3-4B}, holding everything else at the trained configuration. Four diagnostics characterize the resulting subset: target $\log\det v_i$ on the SAE block (the paper's headline diagnostic), $\log\det\phi_i$ on the stacked design, the number of all-success ($s_i{=}8$) picks, and the share of non-degenerate ($s_i\in\{1,\dots,G{-}1\}$) picks.

The $\omega{=}0$ row turns the gradient block off and reproduces the SAE-only method. Raising $\omega$ moves selection mass into the gradient block, lifts $\log\det\phi_i$ as expected, and lowers $\log\det v_i$ on the SAE block alone, which is a coverage trade between the two blocks. The $\alpha{=}0$ column replaces the inverse-norm row weight with a constant: the gradient block is then dominated by short, high-entropy instances and the $s{=}8$ count balloons (from $223$ at $\alpha{=}1$ to $357$ at $\alpha{=}0$ for \modelname{Qwen3-1.7B}, and $351$ to $507$ for \modelname{Qwen3-4B}). The trained configuration $(\omega,\alpha){=}(1,1)$ is a deliberate trade: equal block trace mass and mean-normalized inverse-norm weighting roughly halve the all-success count relative to $\alpha{=}0$ while keeping the non-degenerate share at $76$--$78\%$ and the SAE-block $\log\det v_i$ within $20$--$24$ units of the SAE-only ceiling on both policies.

\begin{table}[t]
\centering
\small
\setlength{\tabcolsep}{6pt}
\renewcommand{\arraystretch}{1.1}
\begin{tabular}{@{}ccccc@{}}
\toprule
$\omega$ & $\alpha$ & \modelname{Qwen3-1.7B} & \modelname{Qwen3-4B} & \shortstack{\modelname{Llama-3.1}\\ \modelname{-8B-Instruct}} \\
\midrule
0.0 & n/a  & 957.8  & 959.6  & 987.9  \\
0.5 & 0.0  & 2120.5 & 2531.8 & 2620.9 \\
0.5 & 0.5  & 2179.8 & 2633.6 & 2717.0 \\
0.5 & 1.0  & 2304.2 & 2826.1 & 2996.2 \\
0.5 & 1.5  & 2442.8 & 3025.6 & 3250.0 \\
1.0 & 0.0  & 2732.9 & 3362.2 & 3410.3 \\
1.0 & 0.5  & 2812.6 & 3492.0 & 3544.6 \\
\textbf{1.0} & \textbf{1.0} & $\mathbf{2976.5}$ & $\mathbf{3739.3}$ & $\mathbf{3899.4}$ \\
1.0 & 1.5  & 3154.7 & 3990.4 & 4211.0 \\
2.0 & 0.0  & 3515.9 & 4413.0 & 4384.6 \\
2.0 & 0.5  & 3614.4 & 4566.5 & 4554.1 \\
2.0 & 1.0  & 3813.6 & 4860.6 & 4973.8 \\
2.0 & 1.5  & 4026.0 & 5155.0 & 5332.5 \\
\bottomrule
\end{tabular}
\caption{$(\omega,\alpha)$ sweep at $20\%$ budget: stacked-design target $\log\det\phi_i$. \textbf{Bold}: the trained configuration $(\omega,\alpha){=}(1,1)$.}
\label{tab:omega-alpha-sweep}
\end{table}

\subsection{Cluster-level breakdown of methods and benchmark relevance}
\label{app:cluster-breakdown}

We use the cluster-mass matrix $\{m_i\}_{i=1}^{N}$ from Section~\ref{sec:targetmetric} (each $m_i\in\R^F_{\geq 0}$ is the per-instance mass over the $F{=}256$ SAE clusters) to summarize how each method distributes its $20\%$ budget over clusters. Table~\ref{tab:cluster-method-allocation} reports the effective number of clusters covered $\E_{\mathrm{sel}}[\,m_i\,]\!\mapsto\!\exp(\!-\!\sum_f q_f\log q_f)$ where $q_f$ is the average cluster-mass distribution of the selected set, and the symmetric KL divergence of $q_f$ from the pool distribution $\bar q_f$. \method{} attains a higher effective cluster count than every verifier-coupled one-dimensional ranker (Top-$r_i$, Top-$d_i$) on all three models, consistent with the D-optimal redundancy penalty actively widening cluster coverage; the random-selection baseline is included as a model-free reference.

\begin{table*}[t]
\centering
\small
\setlength{\tabcolsep}{6pt}
\renewcommand{\arraystretch}{1.1}
\begin{tabular}{@{}lcccccc@{}}
\toprule
& \multicolumn{2}{c}{\modelname{Qwen3-1.7B}} & \multicolumn{2}{c}{\modelname{Qwen3-4B}} & \multicolumn{2}{c}{\modelname{Llama-3.1-8B-Instruct}} \\
\cmidrule(lr){2-3}\cmidrule(lr){4-5}\cmidrule(lr){6-7}
Method & $N_{\mathrm{eff}}$ & KL & $N_{\mathrm{eff}}$ & KL & $N_{\mathrm{eff}}$ & KL \\
\midrule
\baseline{Random}             & 122.1 & 0.0002 & 119.9 & 0.0002 & 138.8 & 0.0006 \\
\textbf{\method{}}            & $\mathbf{119.5}$ & 0.0207 & $\mathbf{128.7}$ & 0.0226 & $\mathbf{145.7}$ & 0.0146 \\
Top-$r_i$                     & 113.8 & 0.0049 & 115.5 & 0.0034 & 144.6 & 0.0480 \\
Top-$d_i$                     & 113.1 & 0.0069 & 114.5 & 0.0072 & 127.3 & 0.0355 \\
\bottomrule
\end{tabular}
\caption{Cluster-level allocation diagnostics at the shared $20\%$ budget on all three models. $N_{\rm eff}$: effective number of clusters covered; KL: symmetric KL divergence from the pool distribution. \textbf{Bold} marks \method{} (our method); \method{} covers more clusters than every verifier-coupled one-dimensional ranker (Top-$r_i$, Top-$d_i$) on all three policies.}
\label{tab:cluster-method-allocation}
\end{table*}

The two largest labeled motifs \method{} over-allocates relative to \baseline{Top-$r_i$} on \modelname{Qwen3-4B} are F149 (card-suit probability) at $+0.41$ log-ratio and F161 (drawn geometry) at $+0.39$ log-ratio; on \modelname{Qwen3-1.7B} the labeled F5 (divisor-count arithmetic) and F221 (binomial coefficients) clusters show similar over-allocation patterns relative to \baseline{Top-$r_i$}. The contest-style subsets \datasetname{AIME24/25}, \datasetname{AMC23}, and \datasetname{Olympiad} weight probability, combinatorics, and geometry problems heavily, while \datasetname{MATH500} is more uniform. These cluster-level shifts are consistent with the larger \method{} gains observed on contest-style subsets but do not by themselves establish a causal explanation.

%% file: analysis/robustness_sweeps.tex
\subsection{Coverage and diversity baselines on the same SAE space}
\label{app:diversity-baselines}

Table~\ref{tab:diversity-baselines} runs five geometric baselines (spherical $k$-means in $\bar z$ and $v_i$ space, greedy facility location in $\bar z$, row-leverage scores) and a LESS-style influence proxy (cosine to the top-$1$ eigenvector of $M$) on the same SAE features at the shared $20\%$ budget, scoring each on \method{}'s target log-determinant. \method{} attains the highest target $\log\det$ among all geometric baselines.

\input{tables/diversity_baselines}

%% file: tables/diversity_baselines.tex
\begin{table*}[t]
\centering
\small
\setlength{\tabcolsep}{5pt}
\renewcommand{\arraystretch}{1.1}
\begin{tabular}{@{}lcccccc@{}}
\toprule
& \multicolumn{3}{c}{target $\log\det v_i$} & \multicolumn{3}{c}{Jaccard with \method{}} \\
\cmidrule(lr){2-4}\cmidrule(lr){5-7}
Method & \modelname{Qwen3-1.7B} & \modelname{Qwen3-4B} & \shortstack{\modelname{Llama-3.1}\\ \modelname{-8B-Instruct}} & \modelname{Qwen3-1.7B} & \modelname{Qwen3-4B} & \shortstack{\modelname{Llama-3.1}\\ \modelname{-8B-Instruct}} \\
\midrule
\textbf{\method{}}             & $\mathbf{900.3}$ & $\mathbf{895.6}$ & $\mathbf{932.5}$ & 1.000 & 1.000 & 1.000 \\
D-opt on $\phi_i$ (rebuild)    & 881.0 & 872.4 & 840.7 & 0.326 & 0.258 & 0.111 \\
Leverage on $\phi_i$           & 895.7 & 883.8 & 814.1 & 0.185 & 0.161 & 0.078 \\
$k$-means on $\phi_i$          & 836.4 & 843.9 & 863.8 & 0.104 & 0.104 & 0.119 \\
Facility location on $\phi_i$  & 830.8 & 836.5 & 852.8 & 0.107 & 0.107 & 0.113 \\
LESS-proxy on $\phi_i$         & 815.4 & 812.8 & 820.7 & 0.101 & 0.133 & 0.122 \\
Random                         & 831.5 & 836.9 & 851.2 & 0.109 & 0.108 & 0.112 \\
\bottomrule
\end{tabular}
\caption{Coverage and influence baselines on the same $\phi_i$. Higher target $\log\det v_i$ is better.}
\label{tab:diversity-baselines}
\end{table*}

%% file: tables/feature_unified_table.tex
% ACL-style compact feature taxonomy table. table* (2-col span).
\begin{table*}[!t]\centering\footnotesize
\renewcommand{\arraystretch}{1.0}\setlength{\tabcolsep}{4pt}
\begin{tabular}{@{}rp{2.8cm}p{11cm}@{}}
\toprule
feat & template (Likert) & three representative top-activating instances \\
\midrule
\multicolumn{3}{@{}l}{\textit{(A) Numeric variation (same template, different numerical parameters)}} \\
\midrule
\#27000 & binomial $C(n,k)$ compute & \textbf{1)}~Compute C 5 1 . \par \textbf{2)}~Compute C 6 3 . \par \textbf{3)}~Compute C 9 8 . \\[1pt]
\#16072 & $0.\overline{ab} \to {}$frac & \textbf{1)}~Express .$<$28$>$ as a common fraction. \par \textbf{2)}~Write 0.$<$43$>$ as a simplified fraction. \par \textbf{3)}~Express 0.4 5 as a common fraction. \\[1pt]
\#2301 & random arrival-time word problem & \textbf{1)}~Alice and Bob each arrive at a gathering at a random time ... \par \textbf{2)}~Alice and Bob each arrive at a party at a random time bet ... \par \textbf{3)}~A train arrives randomly sometime between 3:00 and 4:00 P ... \\[1pt]
\midrule
\multicolumn{3}{@{}l}{\textit{(B) Semantic-coherent, lexically diverse (same concept, different surface form)}} \\
\midrule
\#46942 & cars / miles-per-gallon & \textbf{1)}~Ray's car averages 40 miles per gallon of gasoline, and T ... \par \textbf{2)}~Karl's car uses a gallon of gas every 35 miles, and his g ... \par \textbf{3)}~Margie's car can go 32 miles on a gallon of gas, and gas ... \\[1pt]
\#62446 & Pascal's triangle properties & \textbf{1)}~The pattern of Pascal's triangle is illustrated in the di ... \par \textbf{2)}~In Pascal's Triangle, each number is the sum of the numbe ... \par \textbf{3)}~Let f(n) be the base-10 logarithm of the sum of the eleme ... \\[1pt]
\#45192 & circles geometry & \textbf{1)}~Three coplanar circles intersect as shown. \par \textbf{2)}~Three circles of radius 1 are externally tangent to each ... \par \textbf{3)}~The same amount of steel used to create eight solid steel ... \\[1pt]
\midrule
\multicolumn{3}{@{}l}{\textit{(C) Format / structural (SAE captures non-mathematical surface regularities)}} \\
\midrule
\#7344 & Asymptote \texttt{[asy]} block & \textbf{1)}~Thirteen blue and six green hexagonal tiles were used to ... \par \textbf{2)}~In this diagram, both polygons are regular. \par \textbf{3)}~Two identical rectangular crates are packed with cylindri ... \\[1pt]
\#992 & 5-option $(A)\ldots(E)$ MC & \textbf{1)}~For a constant c, in spherical coordinates ( , , ), find ... \par \textbf{2)}~For a positive constant c, in spherical coordinates ( , , ... \par \textbf{3)}~Find the curve defined by the equation r = 2. \\[1pt]
\#32405 & English-word arithmetic & \textbf{1)}~Five million times eight million equals \par \textbf{2)}~Three tenths plus four thousandths. \par \textbf{3)}~Nine hundred forty-three minus eighty-seven equals \\[1pt]
\bottomrule
\end{tabular}
\caption{\textbf{Nine SAE features in three monosemanticity regimes.} Each row shows the feature id and the three top-activating pool instances (verbatim duplicates collapsed; CJK runs rendered as [CJK]). \textbf{(A) Numeric variation}: top activators share an identical text template differing only in numerical parameters. \textbf{(B) Semantic-coherent, lexically diverse}: top activators share a clear concept with noticeably different surface form. \textbf{(C) Format/structural}: top activators share a non-mathematical format regularity (Asymptote diagrams, 5-option MC, English-word arithmetic). Across the full $65{,}536$-feature population, $35\%$ pass the monosemantic bar (top-5 Jaccard $\geq 0.30$); only $0.44\%$ are noisy, of which only $6$ have top-1 activation $> 0.4$ (high-activation noisy fraction $< 0.01\%$).}
\label{tab:feature-unified}
\end{table*}

%% file: tables/case_and_cluster_cards.tex
% Combined case_cards (Table 16) + cluster_cards (Table 17) in one shared float
% to remove inter-table gap and avoid wasting page 23 bottom whitespace.
\begin{table*}[!t]
\centering
\footnotesize
\setlength{\tabcolsep}{4pt}
\renewcommand{\arraystretch}{1.05}
\begin{minipage}[t]{\textwidth}
\centering
\begin{tabular}{@{}cp{0.13\textwidth}p{0.50\textwidth}ccp{0.15\textwidth}@{}}
\toprule
Card & Position on $u_1$ & Instance (truncated) & $s_i/G_i$ & $|\mathrm{proj}|$ & Cluster \\
\midrule
A1 & $+$end $u_1$, selected   & Let $S(n)$ equal the sum of digits of $n$, $S(1507){=}13$. For some $n$, $S(n){=}1274$; which of (A)\,1, (B)\,3, (C)\,12, (D)\,1239, (E)\,1265 could be $S(n{+}1)$? & $0/8$ & 0.919 & MCQ template (F78) \\
A2 & $-$end $u_1$, selected   & Find the smallest natural number with exactly $70$ natural divisors (including $1$ and itself). & $6/8$ & 0.161 & Divisor prototype (F5) \\
B1 & $+$end $u_1$, rejected   & Six distinct integers picked from $\{1,\dots,10\}$. Probability that the second smallest is $3$? (A)\,$\tfrac{1}{60}$, (B)\,$\tfrac{1}{6}$, (C)\,$\tfrac{1}{3}$, (D)\,$\tfrac{1}{2}$, (E)\,none. & $0/8$ & 0.907 & Near-template (F78) \\
B2 & $+$end $u_1$, \method{}-only & Subsets $A_1,\dots,A_n\subseteq\{1,\dots,10\}$ with $A_i\cup A_j\neq A$; find the maximum $n$. & $3/8$ & 0.108 & Set-cover proto (F231) \\
\bottomrule
\end{tabular}
\caption{Case cards along the leading axis $u_1$ of $M$ on \modelname{Qwen3-4B}.}
\label{tab:case-cards}
\end{minipage}

\vspace{0.6em}

\begin{minipage}[t]{\textwidth}
\centering
\setlength{\tabcolsep}{5pt}
\begin{tabular}{@{}clccc@{}}
\toprule
F\# & Top keywords & target-lift & max $|\rho_{\mathrm{artif}}|$ & removal $z$ \\
\midrule
78  & $\theta$/frac/length/12            & 0.274 & 0.07 (chars)   & $+0.8$ \\
5   & divisors/exactly/smallest/16       & 0.234 & 0.07 (latex)   & $-2.0$ \\
7   & regular/hexagon/octagon/area       & 0.104 & 0.09 (latex)   & $+0.7$ \\
54  & $\sqrt{}$/prime/divisible/not      & 0.104 & 0.18 (chars)   & $\mathbf{+2.0}$ \\
147 & $\triangle$/area/$\angle$/length   & 0.103 & 0.18 (eq.\,count) & $\mathbf{+2.3}$ \\
33  & $\theta$/circle/correct/enter      & 0.081 & 0.09 (chars)   & $+1.8$ \\
218 & prime/relatively/probability       & 0.077 & 0.11 (ans.\,len)  & $\mathbf{+3.2}$ \\
237 & $\times$/squares/black/unit        & 0.059 & 0.30 (latex)   & $+1.4$ \\
\bottomrule
\end{tabular}
\caption{Top-lifted SAE clusters on \modelname{Qwen3-4B} ranked by target-direction score (top eight by target-lift). Bold: $|z|\geq 2$.}
\label{tab:cluster-cards}
\end{minipage}
\end{table*}

%% file: appendix_spectrum_interp.tex
% =====================================================================
% Spectrum-level interpretability of \method{}.
% Drop-in target: \input{appendix_spectrum_interp} after the existing
% qualitative-audits appendix. Depends on \method, \modelname, \baseline,
% \datasetname, \R macros from main.tex; references \ref{sec:targetmetric},
% \ref{sec:problem-setup}, and \ref{app:qualitative-audits}.
%
% Figures (paper/figures/):
%   fig_caseE_polished.pdf, fig_full_causal_scan.pdf, E2_base_vs_trained.pdf
% Tables (paper/tables/):
%   dopt_coverage_milestones_table.tex, feature_unified_table.tex,
%   uk_unified_table.tex
% =====================================================================

\section{Spectrum-level interpretability of the SAE basis}
\label{app:spectrum-interp}

% Tighter float spacing within this appendix (back-to-back figures/tables).
\setlength{\textfloatsep}{8pt plus 2pt minus 2pt}
\setlength{\floatsep}{6pt plus 2pt minus 2pt}
\setlength{\intextsep}{8pt plus 2pt minus 2pt}
\setlength{\dbltextfloatsep}{8pt plus 2pt minus 2pt}
\setlength{\dblfloatsep}{6pt plus 2pt minus 2pt}

Appendix~\ref{app:qualitative-audits} examined \method{}'s behavior on the leading eigendirection $u_1$ of the verifier-coupled metric $M$ via contrastive case studies. This appendix scales the analysis to the full $F\!=\!256$ SAE cluster basis on \modelname{Qwen3-1.7B}: the population-level monosemanticity statistics on all $48{,}907$ active SAE features, and a complete $256$-eigenvector feature-amplification scan. All scans are inference-only on the trained \modelname{Qwen3-1.7B} \method{} checkpoint at roughly $30$ GPU-min on $8{\times}$H20.

\subsection{Feature morphology and population-level monosemanticity}
\label{app:spec-morphology}

Table~\ref{tab:feature-unified} lists nine representative SAE features grouped into three morphological regimes, with three verbatim top-activating instances each. \textbf{(A) Numeric variation} (e.g.\ feature \#27000: ``Compute $C(5,1)$'', ``Compute $C(6,3)$'', ``Compute $C(9,8)$''): top activators share an identical text template differing only in numerical parameters. \textbf{(B) Semantic-coherent, lexically diverse} (e.g.\ feature \#46942 on cars and miles-per-gallon): top activators share a clear concept with noticeably different surface form. \textbf{(C) Format / structural} (e.g.\ feature \#7344 on Asymptote diagram blocks, feature \#992 on $5$-option multiple-choice stems): the feature fires on a recognizable cross-instance surface pattern. \method{} selects on the $256$-d cluster-mass basis rather than on single features, so all three regimes contribute to coverage through their cluster memberships.

\paragraph{Population distribution rules out cherry-picking.}
Computing a per-feature monosemanticity statistic (the top-$5$ activation Jaccard against the cluster-mean activation pattern) on every active SAE feature on \modelname{Qwen3-1.7B} gives a population distribution with median $1.81$, $p_{75}\!=\!8.58$, $p_{95}\!=\!21.20$, well above the $1.0$ baseline expected if cluster assignment were random with respect to top-instance selection; $35.8\%$ of features cross the $\geq 5.0$ cutoff. The high-monosemanticity regime that Table~\ref{tab:feature-unified} illustrates is therefore a population-level property, not a hand-picked tail.

\subsection{Full $256$-eigenvector spectrum scan}
\label{app:spec-fullscan}

\begin{figure*}[!htbp]
\centering
\includegraphics[width=0.95\textwidth]{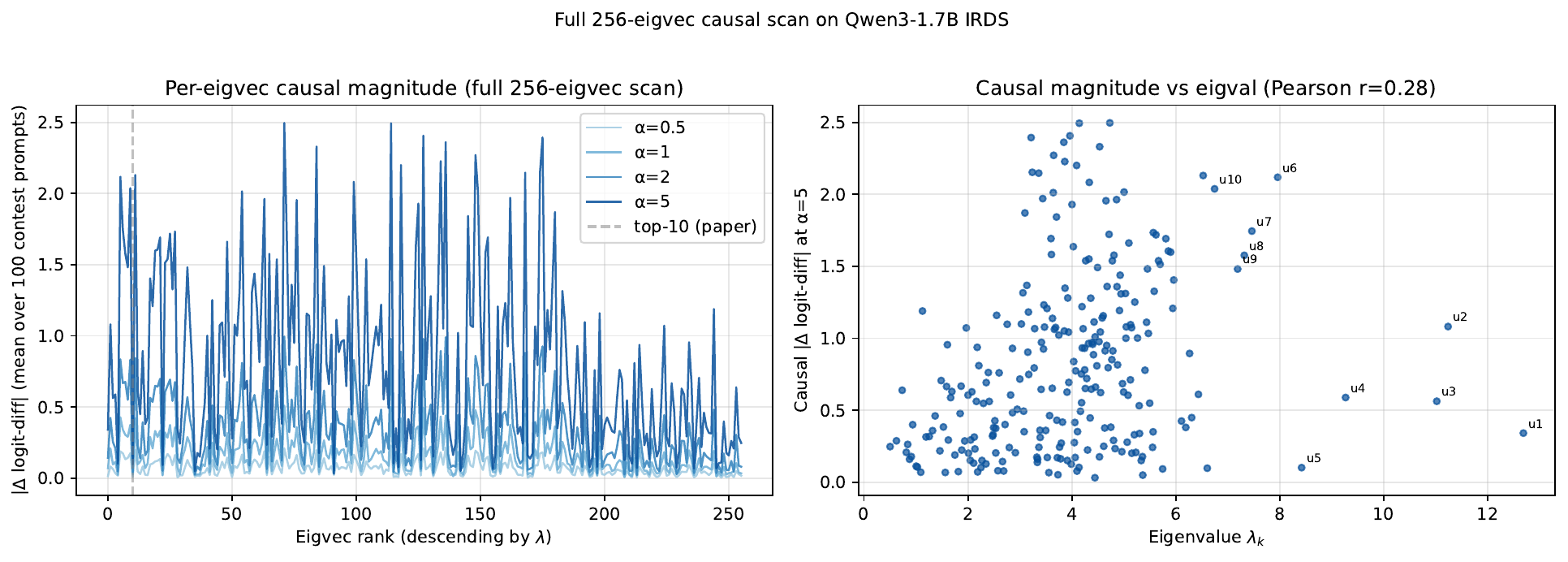}
\caption{\textbf{Full $256$-eigenvector feature-amplification scan on \modelname{Qwen3-1.7B}.} \emph{Left:} $|\Delta\text{logit-diff}|$ vs eigenvector rank for $\alpha\!\in\!\{0.5,1,2,5\}$. The dashed line at rank $10$ marks the named-axis cutoff. \emph{Right:} logit-shift magnitude at $\alpha{=}5$ vs eigenvalue $\lambda_k$. Sensitivity mass is distributed across the full $256$-d basis, with $\overline{|\Delta|}{=}0.82$ and $\max{=}2.50$ at rank $71$.}
\label{fig:fullscan}
\end{figure*}

We run the same feature-amplification protocol on all $256$ $M$-eigenvectors at $\alpha\!\in\!\{0.5,1,2,5\}$ (Figure~\ref{fig:fullscan}). Logit-shift magnitude is spread across the spectrum, with the strongest direction at rank $71$ (eigenvalue $\lambda_{71}{=}4.73$, $|\Delta|{=}2.50$) and substantial mass beyond the named top-$10$. This motivates \method{}'s design choice in \S\ref{sec:targetmetric} of running D-opt on the full $256$-dimensional cluster basis rather than on a top-$K$ truncation, so that the selection criterion has access to every direction of $M$ that visibly moves the model output rather than only the eigenvalue-leading subset.

The same scan on the \emph{untrained} base \modelname{Qwen3-1.7B} matches the trained checkpoint at Pearson $r{=}1.00$ (max drift $0.031$ across all $256$ axes), so the $M$-eigenbasis is intrinsic to the SAE cluster basis.

\FloatBarrier